\documentclass{article}

\usepackage{arxiv}

\usepackage[utf8]{inputenc} 
\usepackage[T1]{fontenc}    
\usepackage{hyperref}       
\usepackage{url}            
\usepackage{booktabs}       
\usepackage{amsfonts}       
\usepackage{nicefrac}       
\usepackage{microtype}      
\usepackage{lipsum}
\usepackage{fancyhdr}       
\usepackage{graphicx}       
\graphicspath{{media/}}     

\usepackage{graphicx}
\usepackage{subcaption}
\usepackage{float}
\usepackage{placeins}
\usepackage{graphicx}
\usepackage{subcaption}
\usepackage{float}
\usepackage{placeins}
\usepackage{amsmath}
\usepackage{array}
\usepackage{graphicx}
\usepackage{subcaption}
\usepackage{float}
\usepackage{placeins}
\usepackage{amsthm}
\newtheorem{definition}{Definition}

\usepackage[utf8]{inputenc} 
\usepackage[T1]{fontenc}    
\usepackage{hyperref}       
\usepackage{url}            
\usepackage{booktabs}       
\usepackage{amsfonts}       
\usepackage{nicefrac}       
\usepackage{microtype}      
\usepackage{xcolor}         
\usepackage{amsmath}
\usepackage{graphicx}
\usepackage{float}
\usepackage{placeins}
\usepackage{multirow}

\title{CORE-RL: Confidence-Oriented Reliability Evaluation of Black-Box Reinforcement Learning Policies}
\author{
  Santhosh GS, Ananya Ravi, Devika Jay, Saurav Prakash, Balaraman Ravindran \\
  Center for Responsible AI \\
  IIT Madras\\
  Chennai, India\\
  \texttt{devika@cerai.in} \\
   \And
  Abhishek Sarakar, Perepu Satheesh Kumar, Kaushik Dey \\
  Ericsson Research \\
  India \\
  \texttt{abhishek.sarkar@ericsson.com} \\
}
\renewcommand{\shorttitle}{\textit{CORE-RL}}
\begin{document}

\maketitle

\begin{abstract}
The deployment of Reinforcement Learning (RL) agents in critical domains must be preceded with a pipeline to evaluate the alignment of the RL agent with complex multi-objective specifications and robustness under real-world environmental drift. However, to protect intellectual property, the RL agent may be delivered for evaluation as opaque executable or remote API, which makes traditional evaluation techniques based on the internals of the policies infeasible. To address this gap, CORE-RL: Confidence-Oriented Reliability Evaluation of black box RL policy is proposed in this paper. The CORE-RL pipeline introduces a Unified Reliability Metric that formally integrates early task termination and safety constraint violations, preventing unsafe policies from masking failures through premature episode halts. By subjecting the policy to a noise certification envelope of perceptual noise, actuation noise and change in environment dynamics, the pipeline computes the finite-sample Clopper-Pearson bounds on unified reliability metric and Hoeffdings' lower bound on reward and safety cost. The pipeline then defines safe operational design domain to report high-confidence certificates for safety and expected performance. Experiments on continuous control tasks demonstrate the CORE-RL pipeline's ability to automatically reject non-compliant policies and map the safe Operational Design Domain (ODD) of safety-aware policies. Thus CORE-RL provides an evaluation framework towards a quantitative, transparent and reproducible, statistical rationale necessary to safely evaluate, compare, and deploy black box RL solutions.
\end{abstract}

\section{Introduction}
The deployment of Reinforcement Learning (RL) in critical domains such as autonomous vehicles, robotics etc has moved from theoretical possibility to practical necessity. However, a critical bottleneck remains in the transition from research to production. The evaluation paradigm has remained largely benchmark-centric: policies are assessed on expected return under nominal test environments. This protocol implicitly assumes access to training details, reward structure, and implementation transparency. However, to protect intellectual property, vendors often deliver policies as opaque executables or remote APIs. Policies are often accessible only through input-output interaction. Their internal representations, optimization history, and reward engineering are opaque.  Under such constraints, rendering traditional white-box verification methods such as formal verification of neural networks or Lyapunov-based stability analysis become inapplicable [1,2]. 

To illustrate the challenge with a real example from industry, with AI-native 6G on the horizon, many radio-network functions such as antenna tilt optimization, uplink/downlink power control, and beamforming are expected to be autonomously executed by RL-driven control policies [17,18]. However, the telecommunications network constitutes critical national infrastructure that supports essential societal services. Consequently, governments and regulatory bodies are increasingly demanding mechanisms to validate and certify radio control functions to ensure safety, trustworthiness, and compliance with regulatory frameworks.
In addition, telecom operators often deploy multi-vendor network stacks, requiring them to evaluate and certify equipment and control policies from multiple vendors to guarantee consistent and optimal network performance. Hence it is important to validate and certify performance and safety of the RL policies through transparent evaluation methodology to ensure equipment safety, and compliance with operational envelopes. Such platforms often integrate multi-vendor instrument stacks (actuators, sensors, robots, controllers), requiring facility teams to evaluate and certify control policies from multiple vendors to guarantee consistent, safe, and optimal experimental throughput. 

In both regulatory and multi-vendor scenarios, direct exposure of the underlying RL policy is typically infeasible due to IP protection and competitive considerations. This limitation becomes particularly acute when policies are delivered as black-box controllers or cloud-hosted APIs and must be evaluated without access to training details [3]. Under these constraints, robust methods for black-box policy evaluation and certification become essential to enable facility operators and oversight bodies to assess safety, reliability, and compliance, thereby providing a practical pathway toward trustworthy and secure deployment of RL in safety critical sectors. 

Although RL agents are typically evaluated using expected return on held-out environments, such metrics provide limited guarantees about safety or reliability under deployment uncertainty. Significant research exists on training safe agents [4] or adversarial testing [5], there is a lack of standardized frameworks that allow a client to formally accept or reject a delivered policy without accessing its source code. Black-box statistical certification constitutes an important but relatively underexplored component of RL evaluation. To bridge this gap, this paper proposes \textsc{CORE-RL}: Confidence-Oriented Reliability Evaluation, a certification pipeline designed for the acceptance testing of third-party RL agents. Unlike standard robustness analysis, which typically assumes white-box access to network weights, or focuses exclusively on adversarial attacks against a single reward scalar, \textsc{CORE-RL} approach is tailored specifically for evaluation of policies without requiring access to proprietary models. The policy is treated purely as an input-output oracle to certify both the performance and strict safety constraint adherence under natural deployment noise.

The key contributions of the paper are:
\begin{itemize}
    \item \textbf{\textsc{CORE-RL} Black-Box RL Evaluation Pipeline:} Formalize black-box RL evaluation as a statistical certification problem that bridges the gap between policy training and real-world deployment. The proposed pipeline provides well-defined evaluation metrics and employs reproducible noise configurations, enabling trustworthy and transparent assessment of black-box RL policies. 
    \item \textbf{Unified Reliability Metric:} For defining system violation, which encompasses both early termination (task failure) and constraint violation (unsafe behavior). The metric provides a formal integration of task performance and safety violations into a single boolean failure condition. This helps in identifying policies that mask safety risks by early failures, in the evaluation pipeline .
   \item \textbf{Operational Design Domain} Statistical certification protocol over a class of noise to formally define the policy's operational design domain for safe and reliable deployment
   \end{itemize}

The remainder of this paper is organized as follows. Section \ref{sec:related_work} reviews related work in Safe RL and policy verification. Section \ref{sec:codeop} details the CORE-RL methodology, formalizing specification and environment deviation and deriving the statistical certificates. Section \ref{sec:experiments} presents our experimental evaluation across multiple continuous control domains. Finally, Section \ref{sec:conclusion} concludes the paper and outlines directions for future research.

\section{Related works}
\label{sec:related_work}
The focus of this paper lies in the intersection of safe reinforcement learning, robust policy certification, and black-box evaluation. This section discusses the related works in these areas, to present the research gaps.

The framework of Constrained Markov Decision Processes (CMDPs) [6] serves as the foundation for most safe RL approaches. To enforce constraints during training, primal-dual methods such as PPO-Lagrangian [4,7] adaptively tune Lagrange multipliers to penalize cost violations. Alternatively, trust-region methods like CPO [8] analytically project policy updates onto a feasible manifold to satisfy safety constraints at every step. While these algorithms are essential for safety-aware policies, they fundamentally differ from the problem of certifying them. As highlighted by [3], policies trained with safety constraints often fail to generalize to the stochasticity of real-world deployment environments.

Formal methods for verifying neural networks often rely on techniques like reachability analysis or barrier certificates to guarantee safety [9]. A significant advancement in this domain is Probably Approximate Safety (PAS) Verification [10], which employs scenario optimization to construct barrier certificates with probabilistic guarantees. PAS formulates verification as a chance-constrained optimization problem, successfully bounding the probability of unsafe transitions. However, such white-box or gray-box methods typically require constructing a mathematical certificate (e.g., a barrier function) or accessing the policy's internal structure. In the cases where vendors protect intellectual property by providing only API access, constructing such certificates is often infeasible. Parallel to safety verification, robustness certification aims to guarantee performance under perturbations. Techniques based on Randomized Smoothing [11] have been adapted to RL to certify policies against adversarial attacks on observations. Recently, ReCePS [12] established state-of-the-art bounds for the cumulative reward of smoothed policies, certifying that performance will not degrade below a specific threshold under $l_p$-norm perturbations. While ReCePS provides rigorous guarantees for \textit{reward}, it does not focus on complementary objective of reliability, specifically, the joint probability of task survival and safety constraint satisfaction. Furthermore, it does not address the epistemic uncertainty of deployment (natural noise). This work attempts to bridge these gaps.

When internal access to a system is denied, Statistical Model Checking (SMC) [13] offers a viable alternative by treating verification as hypothesis testing. SMC bounds the probability of system failure by sampling execution traces. Unlike formal verification methods that prove the absence of errors (which is computationally intractable for black-box neural policies), it is important to quantify the probability of reliability with high statistical confidence, providing a practical ``Acceptance Certificate'' suitable for trustworthy industrial deployment.

\section{CORE-RL: Confidence-Oriented Reliability Evaluation}
\label{sec:codeop}
The problem setting addressed in this paper considers the scenario in which an organization has no visibility into the training objective, the regularization techniques used, and the internal architecture of the RL policy provided by their third-party vendor. The organisation receives a pre-trained policy and the only available interface is a query-response mechanism: the organization provides state observations, and the vendor's API returns actions. The organization evaluates a given black-box policy for its ability to achieve the Key Performance Indexes (KPIs) in which conflicting objectives arise, demanding an optimal trade-off behavior from the RL agent. The section presents the Confidence-Oriented Reliability Evaluation of Opaque RL Policies (CORE-RL) to formalise this evaluation methodology.

 The CORE-RL pipeline evaluates the pre-trained black box policies for deployment-readiness. This is driven by two primary factors:
\begin{enumerate}
    \item \textbf{Specification Deviation:} The vendor may have optimized a surrogate reward function that fails to capture all the client's KPIs, or they may have ignored strict safety constraints to artificially boost apparent performance.
    \item \textbf{Environment Deviation:} Policies trained in deterministic or idealized simulators often fail when subjected to epistemic uncertainty, sensor noise, and actuation friction of real-world deployment. 
\end{enumerate}


CORE-RL evaluates proprietary agents without access to their internal parameters, gradients, or training configurations to certify the policy using.
\begin{enumerate}
    \item \textbf{Unified Reliability Metric} to evaluate the the RL policy for specification deviation by testing the policy under nominal conditions to identify objective (task performance and safety constraints violation) mismatch.
    \item \textbf{Operational Design Domain} to evaluate the the policy's task performance and safety under environment deviation over a class of noise envelopes with reliability and performance certificates.
\end{enumerate}

To formalize the interaction between the evaluator and the RL agent provider, the environments are distinguished and the black-box policy oracle are defined as below:

Let the RL training environment be 
\begin{equation}
\mathcal{M}_{train} = \langle \mathcal{S}, \mathcal{A}, P_{train}, R_{vendor}, C_{vendor}, \gamma \rangle    
\end{equation}
where $R_{vendor}$ and $C_{vendor}$ represent the RL agents internal optimization targets. $P_{train}$ is the state transition probability of the training environment.

It is important to note that $\mathcal{M}_{train}$ is unknown to the deployer who is interested in evaluating the policy before deployment. Let the deployment or evaluation environment, which is different from the training environment, be 
\begin{equation}
\mathcal{M}_{eval} = \langle \mathcal{S}, \mathcal{A}, P_{eval}, R_{client}, C_{client}, \gamma \rangle    
\end{equation}
where $R_{client}$ represents the true KPIs and $C_{client}$ represents the strict safety constraints.$P_{eval}$ is the state transition probability of the evaluation environment.

The agent provider provides a policy $\pi_\theta: \mathcal{S} \to \mathcal{A}$ trained on $\mathcal{M}_{train}$. The evaluator evaluating on  $\mathcal{M}_{eval}$ has access strictly to an oracle $\mathcal{O}(\pi, s_t) \to a_t$. The parameters $\theta$, internal network activations, and the exact formulation of $R_{vendor}$ and $C_{vendor}$ are opaque.

To build trustworthiness and guarantee safety in critical infrastructure, \textsc{CORE-RL} treats the policy's behavior as a stochastic process and bound its failure probability purely through interaction traces on $\mathcal{M}_{eval}$.

\subsection{Evaluating Specification Deviation}
The policy is evaluated by the evaluator for its performance on all KPIs of the environment, which is denoted as $\mathbf{O}_{eval} = [o_1, o_2, \dots, o_n]$. However, the policy may be trained only for a subset of $\mathbf{O}_{eval}$, $\mathbf{O}_{train} = \{o_1, \dots, o_{n-k}\}$, typically prioritizing easily achievable rewards while neglecting complex, multi-objective constraints. Hence the missing KPIs set is to be validated $\mathbf{O}_{miss} = \mathbf{O}_{eval} \setminus \mathbf{O}_{train}$. However, a fragmented definition of failure allows agents to escape the above mentioned level of evaluation. For instance, an agent might achieve zero safety violations simply by terminating episodes immediately, which makes it a safe but useless policy. Conversely, a high-performing agent might complete tasks but routinely violate safety limits.

To do this without white-box inspection, CORE-RL evaluates the agent under zero-noise nominal conditions ($\mathcal{M}_{eval}$ with deterministic sensors). If the agent fails to meet $R_{client}$ or violates $C_{client}$ here, it is immediately rejected for specification deviation. To tighten the evaluation, CORE-RL introduces a \textbf{Unified Reliability Metric}.

\begin{definition}[Unified Reliability Metric]
Let $\tau$ denote a complete episode generated under $\mathcal{M}_{eval}$. 
Let $\text{Goal}(\tau)$ be a client-defined boolean indicator that encodes whether the task objectives are satisfactorily achieved (e.g., goal reached, sustained stability, or KPI thresholds met). 
Let $\text{Violation}(\tau)$ be a boolean indicator defined as:
\[
\text{Violation}(\tau) = \mathbb{I}\left[\sum_{t=0}^{T} C_{client}(s_t, a_t) > d \right],
\]
where $C_{client}(s_t, a_t)$ is the safety constraints imposed by the environment and $d$ is the allowable safety budget.

The episode is declared a Failure if:
\begin{equation}
F(\tau) = \mathbb{I}[\neg \text{Goal}(\tau)] \lor \text{Violation}(\tau)
\label{URM}
\end{equation}
\end{definition}

The Unified Reliability Metric (URM) is defined as 
\begin{equation}
    URM = 1-\frac{1}{N}\sum_{\tau =1} ^N F(\tau)
\end{equation}
$F(\tau)$ indicates the unified failure of the RL agent with respect to task completion and safety. URM value close to 1, indicates that the safety constraints are within the safety budget and the KPIs of the environment are achieved. Thus, the metric provides an direct indicator for accepting or rejecting a policy before actual deployment.   

\subsection{Evaluating Environment Deviation}

Standard evaluation assumes $P_{eval}$ perfectly matches $P_{train}$. However, real-world deployment introduces epistemic uncertainty through sensor noise, actuation friction and change in environment dynamics. To model this without requiring a fully new physics simulator, noise is injected into the true state $s_t$ for evaluation using \textsc{CORE-RL}. Thus, the fully observable Markov property is broken from the perspective of the agent; it is forced to act on a noisy, partial observation $\tilde{s}_t$. Furthermore, the distribution shift from training to evaluation environment is modeled as actuation uncertainty, wherein the action executed by the agent is subjected to dynamic friction or execution errors, resulting in altered state transitions. 

To quantify this, let $s_t \in \mathbb{R}^{d_s}$ be the true environment state and $a_t \in \mathbb{R}^{d_a}$ be the intended action. Let $\Delta_s \in \mathbb{R}^{d_s}$ and $\Delta_a \in \mathbb{R}^{d_a}$ denote the maximum physical ranges (upper bound minus lower bound) of each state and action dimension, respectively. A certification envelope $\mathcal{E}(\sigma_s, \sigma_a)$ is defined by two perturbation components- Perceptual Uncertainty and Actuation Uncertainty. In Perceptual uncertainty, observation provided to the black-box policy is perturbed through a stochastic transformation:
    \begin{equation}
        \tilde{s}_t = f_s(s_t; \sigma_s),
    \end{equation}
    where $f_s$ is a noise-injection function parameterized by the perceptual noise intensity $\sigma_s \in [0,1]$. 
    The function $f_s$ models uncertainty in sensing, estimation, or state measurement and may depend on the operational range of each state dimension.

To simulate actuation uncertainty and distribution shift, the action returned by the policy is transformed prior to environment execution:
    \begin{equation}
        \tilde{a}_t = f_a\big(\pi(\tilde{s}_t); \sigma_a\big),
    \end{equation}
    where $f_a$ is an actuation perturbation function parameterized by noise intensity $\sigma_a \in [0,1]$. 

By evaluating the policy across a dense grid of the noise intensities $(\sigma_s, \sigma_a)$ within $\mathcal{E}$, \textsc{CORE-RL} generates a robustness surface through Statistical Certification Protocol that formally defines the policy's Operational Design Domain (ODD).
\subsubsection{Statistical Certification Protocol}
Statistical certification protocol is formulated as a certification problem by estimating the true failure probability $p_{fail}$ of the black-box policy from finite samples under the noise distribution $\mathcal{E}$. Standard approaches based on the Central Limit Theorem (such as the Wald interval) rely on asymptotic normality. However, as demonstrated by [14], these approximations severely under-cover and exhibit erratic behavior near the boundaries of the probability space ($p \approx 0$ or $p \approx 1$), making them strictly unsafe for critical reliability guarantees. Thus, in this framework, Clopper-Pearson confidence interval [15] is utilised to compute this bound. The Clopper-Pearson guarantee requires that episode-level outcomes 
$\{F(\tau_i)\}_{i=1}^N$ are independent and identically distributed Bernoulli trials. 
To satisfy this assumption, each evaluation episode is initialized from an independently sampled initial-state distribution, and all stochastic perturbations (state and action noise) are independently drawn across episodes. The simulator is reset to a fresh stochastic state at the beginning of each episode, ensuring no carry-over of hidden dynamics or correlated randomness between trials.

With this, the number of failures $k$ in $N$ independent evaluation episodes follows a Binomial distribution $k \sim \text{Bin}(N, p_{fail})$. The Clopper--Pearson method constructs an exact confidence interval by inverting the Binomial cumulative distribution function. This guarantees that the true coverage probability is at least the nominal confidence level $1-\delta$ for all finite sample sizes $N$ and all possible values of $p_{fail}$. This exact finite-sample coverage guarantee makes the Clopper-Pearson interval particularly suitable for black-box safety certification, where collecting extremely large numbers of evaluation episodes is often infeasible.

\subsubsection{Reliability Certificate}
Let $N$ be the number of evaluation episodes collected under noise settings $(\sigma_s, \sigma_a)$. Let $k = \sum_{i=1}^N F(\tau_i)$ be the total count of Failures. For a confidence level $1-\delta$ (e.g., 90\%), the true failure probability $p_{fail}$ is upper bounded by $\hat{p}_{max}$, where:
\begin{equation}
    \hat{p}_{max} = \mathcal{B}_{1-\delta}(k+1, N-k)
\end{equation}
and $\mathcal{B}$ is the quantile function (inverse CDF) of the Beta distribution.

The number of failures $k$ follows a Binomial distribution $Bin(N, p_{fail})$. The Clopper-Pearson upper bound is the value $p$ such that $P(X \le k; p) = \delta$. Utilizing the relationship between the CDF of the Binomial distribution and the regularized incomplete beta function $I_x(a, b)$, specifically $\sum_{j=0}^k \binom{N}{j} p^j (1-p)^{N-j} = I_{1-p}(N-k, k+1)$, allows us to solve for $p$ using the inverse Beta function. This guarantees that $\mathbb{P}(p_{fail} \le \hat{p}_{max}) \ge 1-\delta$.

Determining the number of evaluation episodes required to certify a target reliability threshold is a key practical concern in deployment settings. Appendix~\ref{app:sample_complexity} provides exact finite-sample calculations that quantify the minimum evaluation budget $N$ required to certify a desired reliability level $p^*$ at confidence $1-\delta$. The confidence level quantifies uncertainty in estimating the true failure probability from finite samples, ensuring that the upper bound $\hat{p}_{max}$ satisfies: $ \mathbb{P}(p_{fail} \le \hat{p}_{max}) \ge 1-\delta $.
In contrast, the reliability threshold $p^*$ defines the maximum acceptable failure probability for deployment (e.g., $p^* = 0.05$ corresponds to $95\%$ reliability). 
A policy configuration is certified as deployable only if: $\hat{p}_{max} \le p^*$. Thus, confidence controls statistical uncertainty, whereas the reliability threshold encodes deployment risk tolerance. 

\subsubsection{Performance Certificate}
All KPIs in the environment should be satisfactorily achieved by the agent. However, the agent may or may not be trained for all KPIs in the environment. A performance certificate based on these KPIs (Task completion and safety violation within budget) of the environment is thus essential for evaluation, to ensure minimum performance guarantee when deployed. Let $N$ be the number of evaluation episodes collected under different operating conditions. Let $R_{i}$ be the surrogate reward function defined for $i^{th}$ KPI, defined by the evaluator and $G_i^{k}$ be the cumulative surrogate reward for $i^{th}$ KPI for $k^{th}$ trajectory bounded by [$a_i$,$b_i$]. 

Given sample mean $\hat{\mu}_{G_i}$ and  over $N$ episodes population mean as $\mu_i$ , with 1-$\delta$ confidence, the population mean for the cumulative surrogative reward for $i^{th}$ KPI is 

\begin{equation}
    \mu_i \geq \hat{\mu}_{G_i} - (b_i-a_i)\sqrt{\frac{\ln(1/ \delta)}{2N} }
\end{equation}


The performance for each KPI can be guaranteed by providing a probabilistic lower bound on the expected cumulative surrogate reward of the policy with high confidence and very weak distributional assumptions. By Hoeffding's inequality 
\begin{equation}
Pr( \mu_i \le  \hat{\mu}_{G_i} - \epsilon_i ) \leq \exp \left(-\frac{2N\epsilon_i^2}{(b_i-a_i)^2}\right)
\end{equation}
For 1 - $\delta_i$ confidence, lower bound on the population mean for the $i^{th}$ KPI is
\begin{equation}
    \mu_i \geq \hat{\mu}_{G_i} - (b_i-a_i)\sqrt{\frac{\ln(1/ \delta_i)}{2N} }
\end{equation}
The lower bound provides minimum performance guarantee on the $i^{th}$ KPI. It further shows how the agent performs on each KPI, enabling the evaluator to check whether that KPI was incorporated during training. This formally certifies that, with high confidence, the vendor's policy will deliver an average KPI value of within the Operational Design Domain (ODD). Certifying an Operational Design Domain (ODD) requires simultaneous guarantees over multiple noise configurations. ODD therefore constitutes a family-wise valid high-confidence certified region.

 \section{Experiments}
 \label{sec:experiments}
This section presents empirical validation of the proposed CORE-RL certification pipeline to successfully detect policies that optimize performance while ignoring safety constraints (specification deviation) and that are robust to environmental deviations. The experiments are performed on the Safety Gymnasium benchmark environments [16]. To simulate the epistemic uncertainty inherent in third-party proprietary policy evaluation, independent training runs (random seeds) are treated as distinct APIs of the black box agent. For each environment and algorithm, we trained one policy using a single random seed, seed 0. Once the training was done, we kept that policy fixed and used it as the black-box agent for all the evaluations.
For each trajectory $\tau$, $\mathrm{Success}(\tau)$ indicates successful task completion without premature termination, $c_t$ is the instantaneous safety cost signal, and $d$ is the client-defined cumulative safety budget. The certification pipeline treats fully trained policies as black-box APIs.  Four pre-trained agents, trained on four different algorithms, are evaluated and presented in this section. The details of training the agent are provided in Appendix \ref{App:agents}. 
Code and evaluation artifacts are available in our code base.\footnote{\url{https://anonymous.4open.science/r/core-rl-6393/}}

Table \ref{tab:experiments} summarises the experiments performed to demonstrate the CORE-RL evaluation pipeline. For each agent and evaluation configuration, $N = 10{,}000$ independent episodes are collected to compute statistically tight finite-sample bounds. All certificates are computed at a $90\%$ confidence level.  To ensure a statistically valid region-level guarantee, Bonferroni correction $\delta'$ is applied to allocate a per-configuration confidence level. For each grid point, both the Clopper-Pearson reliability bound and the Hoeffding's inequality performance bound are computed using the adjusted confidence level $1-\delta'$. 
\begin{table}
\caption{Evaluation coverage for 10,000 rollouts per policy--perturbation configuration.}
\label{tab:experiments}
\centering
\resizebox{\textwidth}{!}{%
   \begin{tabular}{lll}
    \toprule
    Environment     & Policy Class     & Environment Deviation\\
     \midrule
SafetyAntVelocity-v1 &
PPO, PPO-Lag, TRPO, TRPO-Lag & Gaussian : $k_a,k_s=\{0, 0.05, 0.1\}$ \\
& & Uniform -$k_a,k_s=\{0, 0.05, 0.1\}$ \\
& & Black-box adversarial $L_\infty$ noise $k_a,k_s=\{0, 0.03, 0.06\}$\\
SafetyPointGoal1-v0 &
PPO, PPO-Lag, TRPO, TRPO-Lag &
Gaussian $k_a,k_s=\{0, 0.05, 0.1\}$\\
& & Uniform $k_a,k_s=\{0, 0.05, 0.1\}$\\
& & Black-box adversarial $L_\infty$ noise $k_a,k_s=\{0, 0.03, 0.06\}$\\
SafetyCarGoal1-v0 &
PPO, PPO-Lag, TRPO, TRPO-Lag &
Gaussian $k_a,k_s=\{0, 0.05, 0.1\}$\\
& & Uniform $k_a,k_s=\{0, 0.05, 0.1\}$\\
& &Black-box adversarial $L_\infty$ noise $k_a,k_s=\{0, 0.03, 0.06\}$\\
& & Change in Environment Dynamics\\ \\
  \bottomrule
   \end{tabular}
   }
 \end{table}

\subsection{Evaluation for Specification Deviation}
The candidate policies are evaluated in a zero-noise environment ($\sigma_s = 0, \sigma_a = 0$) to detect specification deviation. To translate empirical results into deployment decisions, the Clopper--Pearson upper bound on failure probability at $90\%$ confidence is computed. The Clopper Pearson (CP) Upper Bound captures the agents performance on achieving tasks and safety constraints efficiently. A configuration is marked as certified only if the upper bound satisfies $\hat{p}_{max} \le p^* = 0.05$. 
\begin{table}[t]
\caption{Unified Reliability Metric under nominal evaluation at
$d=100$.}
\label{tab:urm_ablation}
\centering
\setlength{\tabcolsep}{2.4pt}
\resizebox{\textwidth}{!}{%
\begin{tabular}{llrrrrrrrl}
\toprule
Environment & Policy
& \multicolumn{2}{c}{Task-only failure}
& \multicolumn{2}{c}{Safety-only failure}
& \multicolumn{2}{c}{Unified failure}
& CORE-RL Evaluation \\
\cmidrule(lr){3-4}\cmidrule(lr){5-6}\cmidrule(lr){7-8} & & Rate & B-CP UB & Rate & B-CP UB & Rate & B-CP UB & \\ \midrule SafetyAnt & PPO & 0.1555 & 0.1640 & 0.9751 & 0.9785 & 0.9913 & 0.9933 & Fail \\ & PPO-Lag & 0.0158 & \textbf{0.0189} & 0.7972 & 0.8063 & 0.8093 & 0.8182 & Fail \\ & TRPO & 0.0146 & \textbf{0.0176} & 0.9978 & 0.9987 & 0.9988 & 0.9994 & Fail\\ & TRPO-Lag & 0.0081 & \textbf{0.0104} & 0.0000 & \textbf{0.0004} & 0.0081 & \textbf{0.0104} & \textbf{Pass} \\ \addlinespace SafetyCar & PPO & 0.0000 & \textbf{0.0004} & 0.1952 & 0.2044 & 0.1952 & 0.2044 & Fail \\ & PPO-Lag & 0.1010 & 0.1081 & 0.0374 & \textbf{0.0420} & 0.1369 & 0.1449 & Fail\\ & TRPO & 0.0000 & \textbf{0.0004} & 0.1970 & 0.2063 & 0.1970 & 0.2063 & Fail\\ & TRPO-Lag & 0.0164 & \textbf{0.0195} & 0.0300 & \textbf{0.0341} & 0.0460 & 0.0510 & Fail \\ \addlinespace SafetyPointGoal & PPO & 0.0000 & \textbf{0.0004} & 0.1060 & 0.1132 & 0.1060 & 0.1132 & Fail\\ & PPO-Lag & 0.0580 & 0.0636 & 0.0202 & \textbf{0.0237} & 0.0775 & 0.0838 & Fail \\ & TRPO & 0.0000 & \textbf{0.0004} & 0.1228 & 0.1305 & 0.1228 & 0.1305 & Fail \\ & TRPO-Lag & 0.0080 & \textbf{0.0103} & 0.0209 & \textbf{0.0244} & 0.0287 & \textbf{0.0328} & \textbf{Pass}\\
\bottomrule
\end{tabular}%
}
\end{table}

For the evaluation in Safety Ant Environment, as shown in Table \ref{tab:urm_ablation}, if agents are evaluated only on the task performance, all agents expect the one trained through PPO exhibit satisfactory performance. However if safety is the only criteria for evaluation, TRPO-Lagrangian trained agent will passes the evaluation pipeline. In Safety Car environment, PPO trained agent cleares the evaluation pipeline if the basis was task performance, but fails when safety alone is considered as the criteria. A similar observation can be found for PPO trained agent in Safety Point Goal environment. Whereas PPO-Lagrangian based agent failes the task only based evaluation but pass the safety only based evaluation. Thus Table \ref{tab:urm_ablation} illustrates the significance of having a Unified Reliability Metric (URM) for considering both task completion and safety performance. Applying the unified failure definition reveals the true deployment risk. 

In Safety Ant and Safety Point Goal Environment, TRPO-Lagrangian trained agent performs well on task completion and safety performance which is indicated by a unified failure with Clopper-Pearson Upper Bound equal to 0.0104, which is within the threshold. Thus passing the CORE-RL Evaluation. However, in Safety Car Environment, the TRPO lagrangian trained agent fails the CORE-RL evaluation pipeline as the Clopper-Pearson Upperbound exceeds the threshold. This indicates that certification before deployment should consider task performance and safety violations, which can be quantified using the Clopper Pearson upper bound on unified failure. This illustrates how CORE-RL enables evaluation by converting raw empirical trajectories into statistically valid deployment decisions.

\subsection{Evaluation for Environment Deviation}
All agents are further evaluated under stochastic environmental perturbations. Although CORE-RL allows general uncertainty operators $f_s$ and $f_a$, without loss of generality, the results are presented for a range-scaled additive noise perturbations of multiple classes. Statistically valid Operational Design Domain (ODD) are generated under family-wise error control. To ensure simultaneous guarantees across all noise configurations, Bonferroni correction is applied over the full grid of evaluated $(\sigma_a, \sigma_s)$ points.  
\begin{table}[t]
\centering
 \caption{Matched SafetyPointGoal Gaussian and Uniform perturbation results at $d=100$. }
\label{tab:point_gaussian_uniform_matched}
\tiny
\setlength{\tabcolsep}{2.3pt}
\resizebox{\textwidth}{!}{%
\begin{tabular}{llrrrrrrrr}
\toprule
Policy & $(\sigma_a,\sigma_s)$
& \multicolumn{4}{c}{Gaussian}
& \multicolumn{4}{c}{Uniform} \\
\cmidrule(lr){3-6}\cmidrule(lr){7-10}
& & Emp.\ fail & B-CP UB & Mean reward & Hoeffding LCB
& Emp.\ fail & B-CP UB & Mean reward & Hoeffding LCB \\
\midrule
PPO & $(0.05,0.00)$ & 0.1090 & 0.1163 & 26.45 & -273.54 & 0.1086 & 0.1159 & 26.49 & -273.50 \\ PPO & $(0.10,0.00)$ & 0.1037 & 0.1109 & 26.34 & -273.65 & 0.1081 & 0.1154 & 26.46 & -273.54 \\ PPO & $(0.10,0.10)$ & 0.4790 & 0.4905 & 2.01 & -297.98 & 0.3221 & 0.3329 & 3.74 & -296.25 \\ PPO-Lag & $(0.05,0.00)$ & 0.0809 & 0.0873 & 11.10 & -288.89 & 0.0821 & 0.0886 & 11.00 & -289.00 \\ PPO-Lag & $(0.10,0.00)$ & 0.0810 & 0.0875 & 11.05 & -288.94 & 0.0800 & 0.0864 & 10.97 & -289.02 \\ PPO-Lag & $(0.10,0.10)$ & 0.7865 & 0.7958 & 0.62 & -299.37 & 0.6472 & 0.6581 & 1.15 & -298.84 \\ TRPO & $(0.05,0.00)$ & 0.1213 & 0.1290 & 26.99 & -273.00 & 0.1202 & 0.1278 & 27.00 & -272.99 \\ TRPO & $(0.10,0.00)$ & 0.1181 & 0.1257 & 26.95 & -273.04 & 0.1212 & 0.1289 & 26.98 & -273.01 \\ TRPO & $(0.10,0.10)$ & 0.3792 & 0.3904 & 2.54 & -297.45 & 0.2799 & 0.2903 & 4.59 & -295.41 \\ TRPO-Lag & $(0.05,0.00)$ & 0.0310 & \textbf{0.0352} & 21.72 & -278.27 & 0.0300 & \textbf{0.0341} & 21.75 & -278.25 \\ TRPO-Lag & $(0.10,0.00)$ & 0.0296 & \textbf{0.0337} & 21.60 & -278.39 & 0.0293 & \textbf{0.0334} & 21.67 & -278.32 \\ TRPO-Lag & $(0.10,0.10)$ & 0.6815 & 0.6921 & 0.95 & -299.04 & 0.4830 & 0.4945 & 1.73 & -298.26 \\
\bottomrule
\end{tabular}
}
 \vspace{-2em}
\end{table}
In Table \ref{tab:point_gaussian_uniform_matched}, the performance of all four agents under perceptual and action noise - Gaussian and Uniform in Safety Point Goal environment- indicate the performance reliability metric based on Hoeffdings' LB and Clopper Pearson Upper Bound. The noise contour that the TRPO Lagrangian performance is within the acceptable limit is [0.05,0.1] for Guassian and Uniform actuation noise. This is known by comparing the Clopper Pearson Upper Bound on unified failure. The Hoeffdings' Lower Bound measures the minimum reward that the agent would receive in the noise envelope. In all other noise configurations, all four agents fail. This defines the noise level tolerance of each agent in the given environment that will help the deployers to define a safe operating region based on agent's robustness to noise.
\begin{figure}[t]
    \centering
    \begin{subfigure}[t]{0.48\textwidth}
        \centering
       \includegraphics[width=1\linewidth,
    trim=0 0 0 78,
    clip]{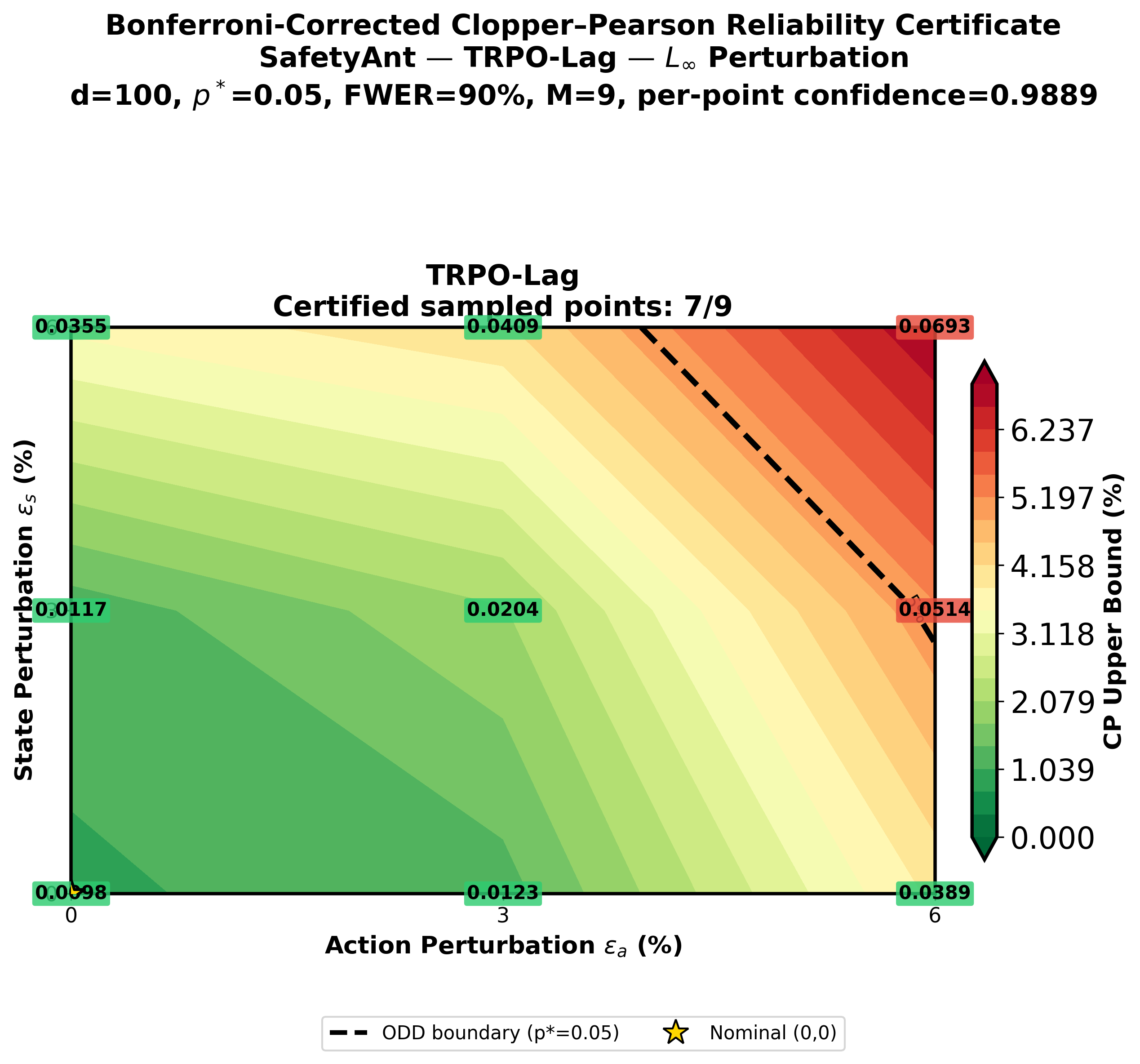}
        \caption{TRPO-Lagrangian}
        \label{fig:safetyant-trpolag-odd}
    \end{subfigure}
    \hfill
    \begin{subfigure}[t]{0.48\textwidth}
        \centering
        \includegraphics[width=1\textwidth,trim=0 0 0 78,
            clip]{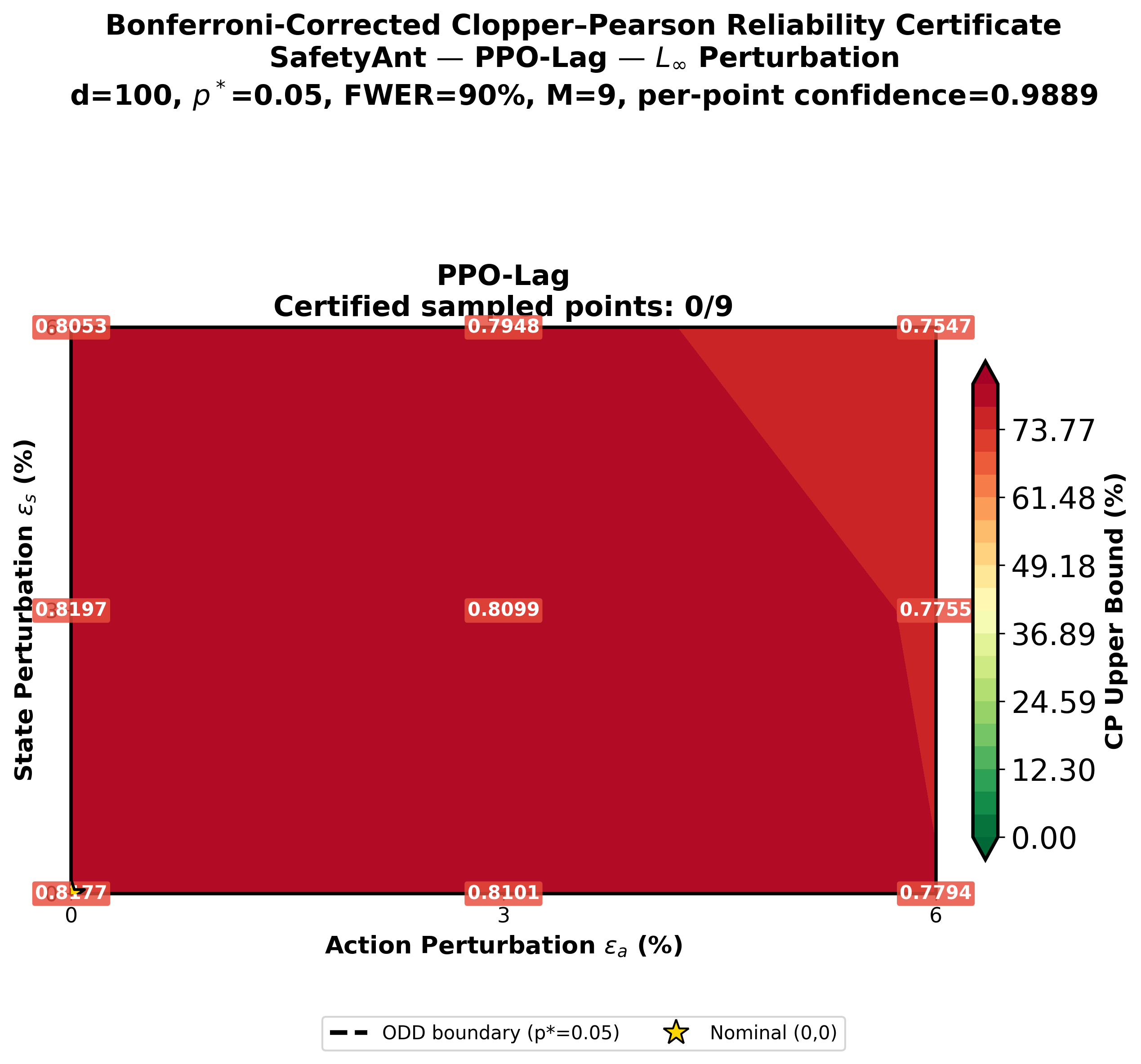}
        \caption{PPO-Lagrangian}
        \label{fig:safetyant-ppolag-odd}
    \end{subfigure}
    \caption{
    Operational Design Domain (ODD) for SafetyAnt under bounded
    adversarial $L_{\infty}$ perturbations at $d=100$ and
    $p^{\star}=0.05$. Gaussian reliability certificates illustrating certified operational design domains (ODDs) for  TRPO-Lag in the left pane attains certification at the stricter threshold d = 100, whereas PPO exhibits a high risk envelope.
    }
    \label{ODD: Safety Ant}
\end{figure}
\begin{figure}[t]
    \centering
    \begin{subfigure}[t]{0.48\textwidth}
        \centering
        \includegraphics[width=1\textwidth,trim=0 0 0 78,
    clip]       {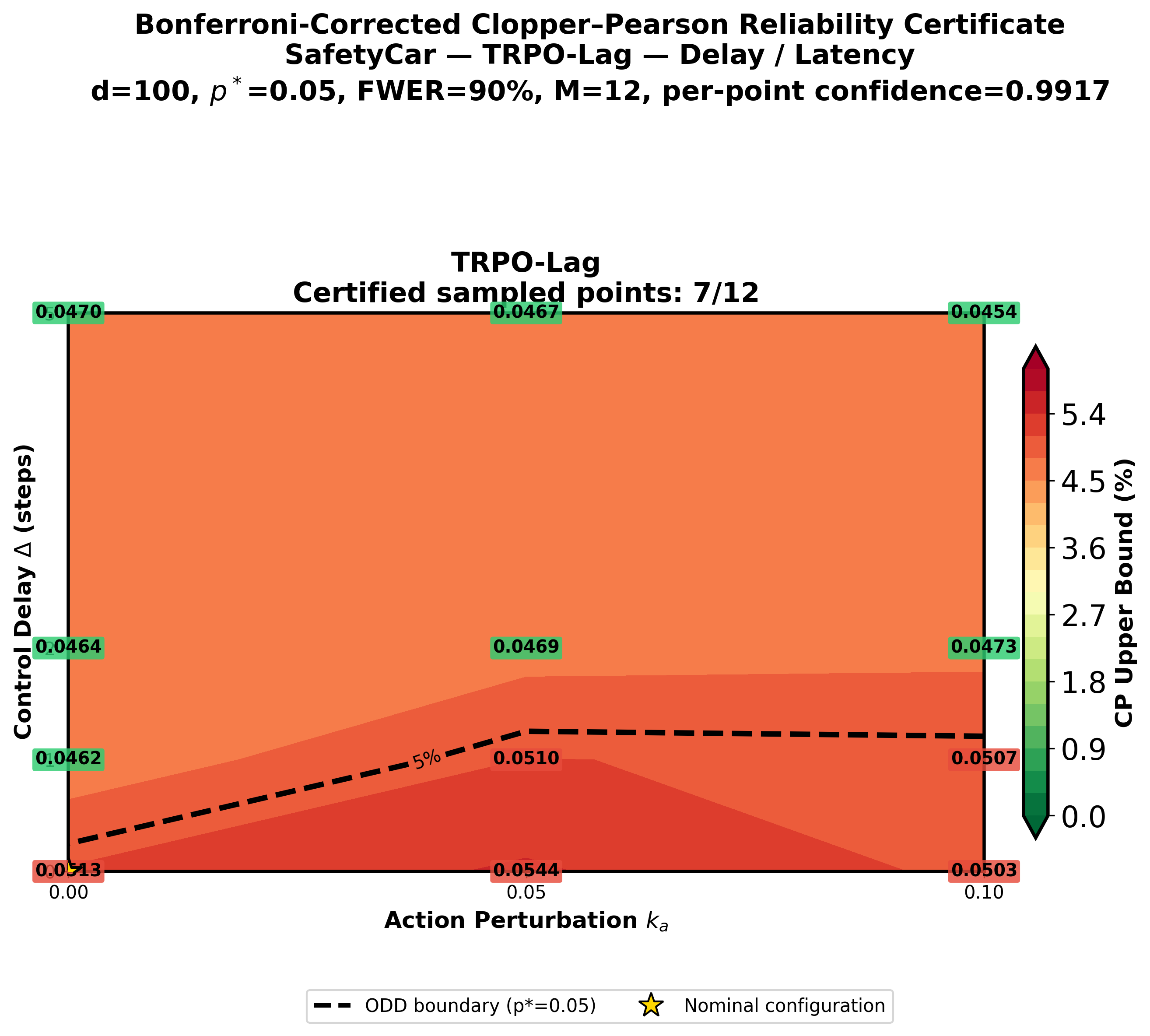}
        \caption{Delay/latency and actuation-noise perturbations}
        \label{fig:safetycar-delay-odd}
    \end{subfigure}
    \hfill
    \begin{subfigure}[t]{0.48\textwidth}
        \centering
        \includegraphics[width=1\textwidth,trim=0 0 0 78,
    clip]        {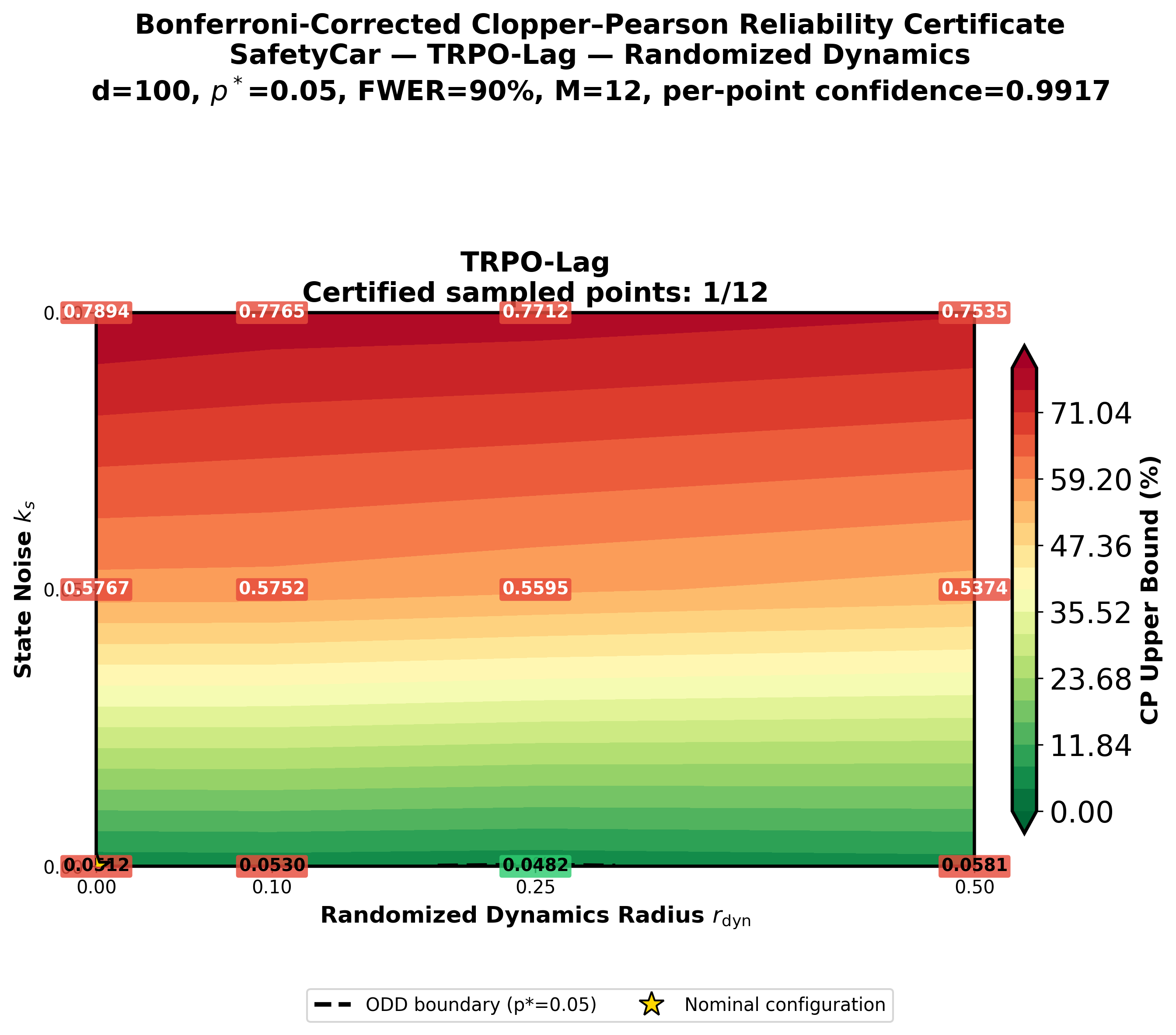}
        \caption{Randomized dynamics and state-noise perturbations}
        \label{fig:safetycar-randdyn-odd}
    \end{subfigure}
    \caption{
    Operational Design Domain (ODD) for the SafetyCar TRPO-Lagrangian
    agent under environment perturbations at $d=100$ and
    $p^{\star}=0.05$.  In each panel, the left subplot
shows the empirical mean reward over 10,000 episodes per perturbation configuration, while the right subplot shows
the Bonferroni-corrected Hoeffding lower bound at 90\% family-wise confidence. Together, these plots show how
task performance changes across the action-noise and state-noise plane, while the lower-bound surface provides a
conservative high-confidence estimate of reward under each perturbation configuration.
    }
    \label{ODD: Safety Car}
\end{figure}
Figure \ref{ODD: Safety Ant}, presents the Operational Design Domain for TRPO-Lagrangian and PPO-Lagrangian under adversarial attack on actuation noise. 
For the bounded black-box $L_\infty$ attack, we perturb the observation seen by the policy and/or the action sent to the environment at every step, while keeping the perturbation magnitude bounded component-wise by $k_s$ for the state and $k_a$ for the action. The perturbations are applied at the policy--environment interface, so the attack requires no access to policy parameters or gradients; robustness is then evaluated over the grid $k_a, k_s \in \{0, 0.03, 0.06\}$. The green color contour in Fig \ref{fig:safetyant-trpolag-odd} represents the safe operation design domain for TRPO Lagrangian based agent in Safety Ant Environment. The ODD for PPO-Lagrangian depicted in Fig \ref{fig:safetyant-ppolag-odd} indicates that a safe operating domain for the agent does not exist under such attack.

The ODD for TRPO-Lagrangian based agent under environment perturbations in Safety Car environment is shown in Figure \ref{ODD: Safety Car}. Four classes of environment perturbation are considered: deterministic physical parameter shifts, randomized dynamics, action delay / control latency and
environment-layout randomization.
The determininstic parameter shift modifies the underlying physical dynamics using mass and friction scaling over $\{0.75,1.00,1.25\}$, with damping fixed at its nominal value. extend this setting by sampling the physical parameters independently for each rollout episode using randomization levels $\rho_{\mathrm{dyn}}\in\{0,0.10,0.25,0.50\}$, capturing episode-to-episode uncertainty in the environment dynamics. Action delay models control latency using $\Delta\in\{0,1,2,5\}$ timesteps, while layout randomization evaluates the policy under nominal, mild, and hard changes to the spatial environment. Fig \ref{fig:safetycar-delay-odd} shows that the TRPO-Lagrangian based agent is not reliable under delay perturbations, where as have a tolerance level on randomised dynamics represented by the green contour in Fig \ref{fig:safetycar-randdyn-odd}. 

Details of the experiments and results on Safety Car, Safety Ant and Safety Point Goal environments are provided in Appendix \ref{App:Car}, \ref{App:Ant}, \ref{App:point} respectively. The Reliability Certificate and ODD plots for all agents under nominal, Gaussian Noise, uniform Noise, adversarial noise and change in environment dynamics indicate the reproducability of the proposed evaluation pipeline. The sensitivity of the reliability certificate to the safety budget of the deployment environment is presented in Appendix \ref{App:Sens}.

Hence the CORE-RL pipeline provides a systematic methodology towards trustworthiness and transparency, to evaluate the performance of the blackbox agent under environment deviations like perceptual, actuation noise and environment deviations. This helps to define the safe operating region under different conditions, so that the performance guarantee is known before actual deployment.

\section{Conclusion}
\label{sec:conclusion}
In this work, CORE-RL - a sample-based certification pipeline for assessing the safe deployment of proprietary RL models was introduced. By treating proprietary policies as black-box APIs, the pipeline enables organizations to independently verify an agent's compliance with safety specifications and its robustness under environmental perturbations. By integrating early task termination and strict safety violations to define a Unified Reliability Metric and ODD, the pipeline could prevent the deployment of policies that can appear acceptable under standard evaluation metrics. The pipeline is statistically grounded, with clear metrics and reproducable noise configurations, making the pipeline trustworthy and transparent. Thus CORE-RL provides a formalised evaluation pipeline that helps in bridinging the gap between training simulators and real-world for safe, trustworthy and reliable deployments. 

 The pipeline can further be scaled up to Multi-Agent Reinforcement Learning (MARL) settings, as most of the real-world deployments implement mulitple blackbox agents interacting with each other. The pipeline also needs to be tested for a real-world deployment setting to understand the implications of the ODD and reliability certificate. This will enable the evaluation of multiple vendor APIs interacting within shared deployment environments. Future work based on this paper is to develop a feedback mechanism based on the findings from the evaluation pipeline to improve the pre-trained policy for improved awareness of the deployment environment constraints.
\section*{References}

\medskip

{
\small

[1] A. Corso, R. Moss, M. Koren, R. Lee, and M. Kochenderfer, “A survey
of algorithms for black-box safety validation of cyber-physical systems,”
Journal of Artificial Intelligence Research, vol. 72, Oct. 2021. \newline
[2] F. Berkenkamp, M. Turchetta, A. P. Schoellig, and A. Krause, “Safe
model-based reinforcement learning with stability guarantees,” 2017. \newline
[3] G. Dulac-Arnold, D. Mankowitz, and T. Hester, “Challenges of real-world
reinforcement learning,” 2019.\newline
[4] A. Stooke, J. Achiam, and P. Abbeel, “Responsive safety in reinforcement
learning by pid lagrangian methods,” 2020.\newline
[5] L. Pinto, J. Davidson, R. Sukthankar, and A. Gupta, “Robust adversarial
reinforcement learning,” 2017.\newline
[6] E. Altman, “Constrained markov decision processes,” 1999.\newline
[7] A. Ray, J. Achiam, and D. Amodei, “Benchmarking safe exploration in
deep reinforcement learning,”\newline
[8] J. Achiam, D. Held, A. Tamar, and P. Abbeel, “Constrained policy
optimization,” 2017.\newline
[9] A. D. Ames, X. Xu, J. W. Grizzle, and P. Tabuada, “Control barrier
function based quadratic programs for safety critical systems,” IEEE
Transactions on Automatic Control, vol. 62, no. 8, pp. 3861–3876, 2017.\newline
[10] A. J. Singh and A. Easwaran, “Pas: Probably approximate safety verifi-
cation of reinforcement learning policy using scenario optimization,” in
Proceedings of the 23rd International Conference on Autonomous Agents
and Multiagent Systems, AAMAS ’24, (Richland, SC), p. 1745–1753,
International Foundation for Autonomous Agents and Multiagent Systems,
2024.\newline
[11] J. M. Cohen, E. Rosenfeld, and J. Z. Kolter, “Certified adversarial
robustness via randomized smoothing,” 2019.\newline
[12] R. Mu, L. S. Marcolino, T. Zhang, Y. Zhang, X. Huang, and W. Ruan,
“Reward certification for policy smoothed reinforcement learning,” 2023.\newline
[13] A. Legay, A. Lukina, L. M. Traonouez, J. Yang, S. A. Smolka, and
R. Grosu, Statistical Model Checking, p. 478–504. Berlin, Heidelberg:
Springer-Verlag, 2022.\newline
[14] L. D. Brown, T. T. Cai, and A. DasGupta, “Interval estimation for a
binomial proportion,” Statistical Science, vol. 16, no. 2, pp. 101–117,
2001. \newline
[15] C. J. CLOPPER and E. S. PEARSON, “The use of confidence or fiducial
limits illustrated in the case of the binomial,” Biometrika, vol. 26, pp. 404–
413, 12 1934. \newline
[16] Ji, Jiaming, Borong Zhang, Jiayi Zhou, Xuehai Pan, Weidong Huang, Ruiyang Sun, Yiran Geng, Yifan Zhong, Josef Dai, and Yaodong Yang. "Safety gymnasium: A unified safe reinforcement learning benchmark." Advances in Neural Information Processing Systems 36 (2023): 18964-18993. \newline
[17] Guo, Shuaishuai, et al. "Towards transparent 6g ai-ran: A survey on explainable deep reinforcement learning for intelligent network slicing." Journal of Information and Intelligence (2025). \newline
[18] Feng, John. "Deep Reinforcement Learning–Based Beamforming for RIS-Aided 6G Systems." Journal of Advanced Digital Communications 2.1 (2025): 3.\newline
}


\appendix
\raggedbottom

\setlength{\textfloatsep}{8pt}
\setlength{\floatsep}{8pt}
\setlength{\intextsep}{8pt}
\section{Training Details and Compute Resources}
\label{App:agents}

All policies were trained on the corresponding Safety-Gymnasium environments
using \textsc{OmniSafe} v0.5.0. Training was performed for $10^7$
environment interactions over 500 epochs, with 20,000 steps per epoch.
Actor and critic networks use two hidden layers of $[64,64]$ units with
$\tanh$ activations. We use $\gamma=\gamma_c=0.99$ and GAE with
$\lambda=\lambda_c=0.95$. Observations are normalized during training.
All policy training and rollout-based evaluation experiments were performed
on a compute node equipped with two NVIDIA A100 GPUs. Individual training
and rollout jobs were assigned to a single GPU, while multiple experiments
were distributed across the two GPUs to parallelize the overall workload.

After training, all policy parameters
were fixed for the subsequent CORE-RL evaluations.

\subsection{PPO Based Agent}

PPO uses the clipped surrogate objective with clipping ratio $0.2$,
target KL divergence $0.02$, and batch size 64. The actor and critic
learning rates are $3\times10^{-4}$.

\subsection{PPO-Lagrangian Based Agent}

PPO-Lagrangian augments PPO with a Lagrangian safety-cost constraint.
The training cost limit is $25$, with initial multiplier $0.001$ and
multiplier learning rate $0.035$ using Adam.

\subsection{TRPO Based Agent}

TRPO uses KL-constrained trust-region updates with target KL divergence
$0.01$, batch size 128, 15 conjugate-gradient iterations, and damping
coefficient $0.1$.

\subsection{TRPO-Lagrangian Based Agent}

TRPO-Lagrangian combines the TRPO trust-region update with the same
Lagrangian safety-cost constraint, using cost limit $25$, initial
multiplier $0.001$, and multiplier learning rate $0.035$.
\begingroup
\small
\section{Complete Results for Safety Car}
\label{App:Car}

\renewcommand\thefigure{\thesection.\arabic{figure}} 
\renewcommand\thetable{\thesection.\arabic{table}}
\subsection{Gaussian Noise}

The Gaussian perturbation experiments evaluate state and action noise over
$k_a,k_s\in\{0.00,0.05,0.10\}$. Each policy--configuration pair is
evaluated using $N=10{,}000$ episodes. Certification uses a Bonferroni
family size of $M=9$ per policy, fixed safety budget $d=100$, and
deployment failure threshold $p^\star=0.05$.










\begin{table}[!htbp]
\centering
\caption{Reliability certificates for PPO on SafetyCar under Gaussian perturbations.}
\label{tab:car-gaussian-ppo-rel}

\small
\renewcommand{\arraystretch}{1.12}
\setlength{\tabcolsep}{5pt}

\begin{tabular*}{\textwidth}{
@{\extracolsep{\fill}}
l c r r r r c
}
\toprule
\textbf{Policy}
& \textbf{$(k_a,k_s)$}
& \textbf{$\hat p_{\mathrm{task}}$}
& \textbf{$\hat p_{\mathrm{safety}}$}
& \textbf{$\hat p_F$}
& \textbf{B-CP UB}
& \textbf{Decision} \\
\midrule

PPO & $(0.00,0.00)$ & 0.0000 & 0.1952 & 0.1952 & 0.2044 & FAIL \\
PPO & $(0.00,0.05)$ & 0.7014 & 0.1462 & 0.7450 & 0.7549 & FAIL \\
PPO & $(0.00,0.10)$ & 0.8066 & 0.1379 & 0.8372 & 0.8456 & FAIL \\

\addlinespace[2pt]

PPO & $(0.05,0.00)$ & 0.0000 & 0.1960 & 0.1960 & 0.2052 & FAIL \\
PPO & $(0.05,0.05)$ & 0.7067 & 0.1496 & 0.7474 & 0.7573 & FAIL \\
PPO & $(0.05,0.10)$ & 0.7984 & 0.1367 & 0.8272 & 0.8358 & FAIL \\

\addlinespace[2pt]

PPO & $(0.10,0.00)$ & 0.0000 & 0.2037 & 0.2037 & 0.2131 & FAIL \\
PPO & $(0.10,0.05)$ & 0.7081 & 0.1442 & 0.7512 & 0.7610 & FAIL \\
PPO & $(0.10,0.10)$ & 0.8063 & 0.1393 & 0.8334 & 0.8418 & FAIL \\

\bottomrule
\end{tabular*}
\end{table}










\begin{table}[!htbp]
\centering
\caption{Reliability certificates for PPO-Lagrangian on SafetyCar under Gaussian perturbations.}
\label{tab:car-gaussian-ppolag-rel}

\small
\renewcommand{\arraystretch}{1.12}
\setlength{\tabcolsep}{5pt}

\begin{tabular*}{\textwidth}{
@{\extracolsep{\fill}}
l c r r r r c
}
\toprule
\textbf{Policy}
& \textbf{$(k_a,k_s)$}
& \textbf{$\hat p_{\mathrm{task}}$}
& \textbf{$\hat p_{\mathrm{safety}}$}
& \textbf{$\hat p_F$}
& \textbf{B-CP UB}
& \textbf{Decision} \\
\midrule

PPO-Lag & $(0.00,0.00)$ & 0.1010 & 0.0374 & 0.1369 & 0.1449 & FAIL \\
PPO-Lag & $(0.00,0.05)$ & 0.8307 & 0.1846 & 0.8573 & 0.8652 & FAIL \\
PPO-Lag & $(0.00,0.10)$ & 0.8776 & 0.2035 & 0.9002 & 0.9069 & FAIL \\

\addlinespace[2pt]

PPO-Lag & $(0.05,0.00)$ & 0.1018 & 0.0400 & 0.1396 & 0.1477 & FAIL \\
PPO-Lag & $(0.05,0.05)$ & 0.8259 & 0.1888 & 0.8545 & 0.8625 & FAIL \\
PPO-Lag & $(0.05,0.10)$ & 0.8749 & 0.2009 & 0.8965 & 0.9034 & FAIL \\

\addlinespace[2pt]

PPO-Lag & $(0.10,0.00)$ & 0.1003 & 0.0415 & 0.1401 & 0.1482 & FAIL \\
PPO-Lag & $(0.10,0.05)$ & 0.8288 & 0.1888 & 0.8563 & 0.8642 & FAIL \\
PPO-Lag & $(0.10,0.10)$ & 0.8731 & 0.2017 & 0.8958 & 0.9027 & FAIL \\

\bottomrule
\end{tabular*}
\end{table}










\begin{table}[!htbp]
\centering
\caption{Reliability certificates for TRPO on SafetyCar under Gaussian perturbations.}
\label{tab:car-gaussian-trpo-rel}

\small
\renewcommand{\arraystretch}{1.12}
\setlength{\tabcolsep}{5pt}

\begin{tabular*}{\textwidth}{
@{\extracolsep{\fill}}
l c r r r r c
}
\toprule
\textbf{Policy}
& \textbf{$(k_a,k_s)$}
& \textbf{$\hat p_{\mathrm{task}}$}
& \textbf{$\hat p_{\mathrm{safety}}$}
& \textbf{$\hat p_F$}
& \textbf{B-CP UB}
& \textbf{Decision} \\
\midrule

TRPO & $(0.00,0.00)$ & 0.0000 & 0.1970 & 0.1970 & 0.2063 & FAIL \\
TRPO & $(0.00,0.05)$ & 0.7137 & 0.2053 & 0.7672 & 0.7768 & FAIL \\
TRPO & $(0.00,0.10)$ & 0.8285 & 0.2199 & 0.8610 & 0.8688 & FAIL \\

\addlinespace[2pt]

TRPO & $(0.05,0.00)$ & 0.0000 & 0.2015 & 0.2015 & 0.2108 & FAIL \\
TRPO & $(0.05,0.05)$ & 0.7153 & 0.2091 & 0.7697 & 0.7793 & FAIL \\
TRPO & $(0.05,0.10)$ & 0.8293 & 0.2163 & 0.8605 & 0.8683 & FAIL \\

\addlinespace[2pt]

TRPO & $(0.10,0.00)$ & 0.0000 & 0.1938 & 0.1938 & 0.2030 & FAIL \\
TRPO & $(0.10,0.05)$ & 0.7185 & 0.2122 & 0.7739 & 0.7834 & FAIL \\
TRPO & $(0.10,0.10)$ & 0.8327 & 0.2276 & 0.8650 & 0.8727 & FAIL \\

\bottomrule
\end{tabular*}
\end{table}

\begin{table}[!htbp]
\centering
\caption{Performance and cost certificates for TRPO-Lagrangian on SafetyCar under Gaussian perturbations.}
\label{tab:car-gaussian-trpolag-perf}

\small
\renewcommand{\arraystretch}{1.12}
\setlength{\tabcolsep}{5pt}

\begin{tabular*}{\textwidth}{
@{\extracolsep{\fill}}
l c r r r r
}
\toprule
\textbf{Policy}
& \textbf{$(k_a,k_s)$}
& \textbf{$\bar{R}$}
& \textbf{$R_{\mathrm{H\text{-}LCB}}$}
& \textbf{$\bar{C}$}
& \textbf{$C_{\mathrm{H\text{-}UCB}}$} \\
\midrule

TRPO-Lag & $(0.00,0.00)$ & 23.35 & -276.64 & 28.88 & 43.88 \\
TRPO-Lag & $(0.00,0.05)$ & -0.82 & -300.81 & 37.04 & 52.04 \\
TRPO-Lag & $(0.00,0.10)$ & -1.54 & -301.53 & 37.38 & 52.38 \\

\addlinespace[2pt]

TRPO-Lag & $(0.05,0.00)$ & 23.61 & -276.38 & 29.39 & 44.39 \\
TRPO-Lag & $(0.05,0.05)$ & -0.82 & -300.81 & 35.39 & 50.39 \\
TRPO-Lag & $(0.05,0.10)$ & -1.56 & -301.56 & 36.17 & 51.17 \\

\addlinespace[2pt]

TRPO-Lag & $(0.10,0.00)$ & 24.12 & -275.87 & 28.63 & 43.63 \\
TRPO-Lag & $(0.10,0.05)$ & -0.87 & -300.86 & 36.23 & 51.23 \\
TRPO-Lag & $(0.10,0.10)$ & -1.56 & -301.56 & 37.35 & 52.35 \\

\bottomrule
\end{tabular*}
\end{table}

\begin{table}[!htbp]
\centering
\caption{Reliability certificates for TRPO-Lagrangian on SafetyCar under Gaussian perturbations.}
\label{tab:car-gaussian-trpolag-rel}

\small
\renewcommand{\arraystretch}{1.12}
\setlength{\tabcolsep}{5pt}

\begin{tabular*}{\textwidth}{
@{\extracolsep{\fill}}
l c r r r r c
}
\toprule
\textbf{Policy}
& \textbf{$(k_a,k_s)$}
& \textbf{$\hat p_{\mathrm{task}}$}
& \textbf{$\hat p_{\mathrm{safety}}$}
& \textbf{$\hat p_F$}
& \textbf{B-CP UB}
& \textbf{Decision} \\
\midrule

TRPO-Lag & $(0.00,0.00)$ & 0.0164 & 0.0300 & 0.0460 & 0.0510 & FAIL \\
TRPO-Lag & $(0.00,0.05)$ & 0.6921 & 0.1417 & 0.7406 & 0.7506 & FAIL \\
TRPO-Lag & $(0.00,0.10)$ & 0.8014 & 0.1424 & 0.8338 & 0.8422 & FAIL \\

\addlinespace[2pt]

TRPO-Lag & $(0.05,0.00)$ & 0.0161 & 0.0288 & 0.0444 & \textbf{0.0493} & \textbf{PASS} \\
TRPO-Lag & $(0.05,0.05)$ & 0.6971 & 0.1325 & 0.7434 & 0.7533 & FAIL \\
TRPO-Lag & $(0.05,0.10)$ & 0.8124 & 0.1308 & 0.8418 & 0.8501 & FAIL \\

\addlinespace[2pt]

TRPO-Lag & $(0.10,0.00)$ & 0.0154 & 0.0286 & 0.0435 & \textbf{0.0484} & \textbf{PASS} \\
TRPO-Lag & $(0.10,0.05)$ & 0.7042 & 0.1339 & 0.7480 & 0.7579 & FAIL \\
TRPO-Lag & $(0.10,0.10)$ & 0.8116 & 0.1422 & 0.8420 & 0.8503 & FAIL \\

\bottomrule
\end{tabular*}
\end{table}

\FloatBarrier

\subsection{Uniform Noise}

The Uniform perturbation experiments evaluate state and action noise over
$k_a,k_s\in\{0.00,0.05,0.10\}$. Each policy--configuration pair is
evaluated using $N=10{,}000$ episodes. Certification uses a Bonferroni
family size of $M=9$ per policy, fixed safety budget $d=100$, and
deployment failure threshold $p^\star=0.05$.




\begin{table}[!htbp]
\centering
\caption{Reliability certificates for PPO on SafetyCar under Uniform perturbations.}
\label{tab:car-uniform-ppo-rel}
\small
\renewcommand{\arraystretch}{1.12}
\setlength{\tabcolsep}{5pt}

\begin{tabular*}{\textwidth}{@{\extracolsep{\fill}} l c r r r r c}
\toprule
\textbf{Policy}
& \textbf{$(k_a,k_s)$}
& \textbf{$\hat p_{\mathrm{task}}$}
& \textbf{$\hat p_{\mathrm{safety}}$}
& \textbf{$\hat p_F$}
& \textbf{B-CP UB}
& \textbf{Decision} \\
\midrule
PPO & $(0.00,0.00)$ & 0.0000 & 0.1952 & 0.1952 & 0.2044 & FAIL \\
PPO & $(0.00,0.05)$ & 0.5885 & 0.1534 & 0.6481 & 0.6590 & FAIL \\
PPO & $(0.00,0.10)$ & 0.7327 & 0.1455 & 0.7738 & 0.7833 & FAIL \\
PPO & $(0.05,0.00)$ & 0.0001 & 0.1997 & 0.1998 & 0.2091 & FAIL \\
PPO & $(0.05,0.05)$ & 0.5858 & 0.1540 & 0.6433 & 0.6542 & FAIL \\
PPO & $(0.05,0.10)$ & 0.7372 & 0.1447 & 0.7760 & 0.7855 & FAIL \\
PPO & $(0.10,0.00)$ & 0.0000 & 0.1966 & 0.1966 & 0.2059 & FAIL \\
PPO & $(0.10,0.05)$ & 0.5925 & 0.1542 & 0.6477 & 0.6586 & FAIL \\
PPO & $(0.10,0.10)$ & 0.7370 & 0.1464 & 0.7739 & 0.7834 & FAIL \\
\bottomrule
\end{tabular*}
\end{table}





\begin{table}[!htbp]
\centering
\caption{Reliability certificates for PPO-Lagrangian on SafetyCar under Uniform perturbations.}
\label{tab:car-uniform-ppolag-rel}
\small
\renewcommand{\arraystretch}{1.12}
\setlength{\tabcolsep}{5pt}

\begin{tabular*}{\textwidth}{@{\extracolsep{\fill}} l c r r r r c}
\toprule
\textbf{Policy}
& \textbf{$(k_a,k_s)$}
& \textbf{$\hat p_{\mathrm{task}}$}
& \textbf{$\hat p_{\mathrm{safety}}$}
& \textbf{$\hat p_F$}
& \textbf{B-CP UB}
& \textbf{Decision} \\
\midrule
PPO-Lag & $(0.00,0.00)$ & 0.1010 & 0.0374 & 0.1369 & 0.1449 & FAIL \\
PPO-Lag & $(0.00,0.05)$ & 0.7376 & 0.1469 & 0.7723 & 0.7818 & FAIL \\
PPO-Lag & $(0.00,0.10)$ & 0.8406 & 0.1853 & 0.8654 & 0.8731 & FAIL \\
PPO-Lag & $(0.05,0.00)$ & 0.1046 & 0.0415 & 0.1445 & 0.1527 & FAIL \\
PPO-Lag & $(0.05,0.05)$ & 0.7427 & 0.1486 & 0.7772 & 0.7867 & FAIL \\
PPO-Lag & $(0.05,0.10)$ & 0.8411 & 0.1850 & 0.8679 & 0.8756 & FAIL \\
PPO-Lag & $(0.10,0.00)$ & 0.1025 & 0.0397 & 0.1409 & 0.1490 & FAIL \\
PPO-Lag & $(0.10,0.05)$ & 0.7353 & 0.1484 & 0.7714 & 0.7809 & FAIL \\
PPO-Lag & $(0.10,0.10)$ & 0.8440 & 0.1909 & 0.8701 & 0.8777 & FAIL \\
\bottomrule
\end{tabular*}
\end{table}





\begin{table}[!htbp]
\centering
\caption{Reliability certificates for TRPO on SafetyCar under Uniform perturbations.}
\label{tab:car-uniform-trpo-rel}
\small
\renewcommand{\arraystretch}{1.12}
\setlength{\tabcolsep}{5pt}

\begin{tabular*}{\textwidth}{@{\extracolsep{\fill}} l c r r r r c}
\toprule
\textbf{Policy}
& \textbf{$(k_a,k_s)$}
& \textbf{$\hat p_{\mathrm{task}}$}
& \textbf{$\hat p_{\mathrm{safety}}$}
& \textbf{$\hat p_F$}
& \textbf{B-CP UB}
& \textbf{Decision} \\
\midrule
TRPO & $(0.00,0.00)$ & 0.0000 & 0.1970 & 0.1970 & 0.2063 & FAIL \\
TRPO & $(0.00,0.05)$ & 0.4997 & 0.1943 & 0.5936 & 0.6048 & FAIL \\
TRPO & $(0.00,0.10)$ & 0.7577 & 0.2233 & 0.8073 & 0.8162 & FAIL \\
TRPO & $(0.05,0.00)$ & 0.0000 & 0.1976 & 0.1976 & 0.2069 & FAIL \\
TRPO & $(0.05,0.05)$ & 0.5015 & 0.1956 & 0.5991 & 0.6103 & FAIL \\
TRPO & $(0.05,0.10)$ & 0.7562 & 0.2098 & 0.8000 & 0.8091 & FAIL \\
TRPO & $(0.10,0.00)$ & 0.0000 & 0.1968 & 0.1968 & 0.2061 & FAIL \\
TRPO & $(0.10,0.05)$ & 0.5013 & 0.1850 & 0.5917 & 0.6030 & FAIL \\
TRPO & $(0.10,0.10)$ & 0.7560 & 0.2147 & 0.7988 & 0.8079 & FAIL \\
\bottomrule
\end{tabular*}
\end{table}





\begin{table}[!htbp]
\centering
\caption{Reliability certificates for TRPO-Lagrangian on SafetyCar under Uniform perturbations.}
\label{tab:car-uniform-trpolag-rel}
\small
\renewcommand{\arraystretch}{1.12}
\setlength{\tabcolsep}{5pt}

\begin{tabular*}{\textwidth}{@{\extracolsep{\fill}} l c r r r r c}
\toprule
\textbf{Policy}
& \textbf{$(k_a,k_s)$}
& \textbf{$\hat p_{\mathrm{task}}$}
& \textbf{$\hat p_{\mathrm{safety}}$}
& \textbf{$\hat p_F$}
& \textbf{B-CP UB}
& \textbf{Decision} \\
\midrule
TRPO-Lag & $(0.00,0.00)$ & 0.0164 & 0.0300 & 0.0460 & 0.0510 & FAIL \\
TRPO-Lag & $(0.00,0.05)$ & 0.4989 & 0.1165 & 0.5620 & 0.5734 & FAIL \\
TRPO-Lag & $(0.00,0.10)$ & 0.7289 & 0.1378 & 0.7705 & 0.7801 & FAIL \\
TRPO-Lag & $(0.05,0.00)$ & 0.0182 & 0.0315 & 0.0490 & 0.0542 & FAIL \\
TRPO-Lag & $(0.05,0.05)$ & 0.5022 & 0.1167 & 0.5670 & 0.5784 & FAIL \\
TRPO-Lag & $(0.05,0.10)$ & 0.7308 & 0.1332 & 0.7713 & 0.7808 & FAIL \\
TRPO-Lag & $(0.10,0.00)$ & 0.0148 & 0.0311 & 0.0451 & 0.0501 & FAIL \\
TRPO-Lag & $(0.10,0.05)$ & 0.5080 & 0.1182 & 0.5713 & 0.5826 & FAIL \\
TRPO-Lag & $(0.10,0.10)$ & 0.7322 & 0.1365 & 0.7728 & 0.7823 & FAIL \\
\bottomrule
\end{tabular*}
\end{table}

\FloatBarrier






\subsection{Bounded Adversarial $L_\infty$ Noise}

The bounded $L_\infty$ perturbation experiments evaluate state and action
perturbations over $k_a,k_s\in\{0.00,0.03,0.06\}$. Each
policy--configuration pair is evaluated using $N=10{,}000$ episodes.
Certification uses a Bonferroni family size of $M=9$ per policy, fixed
safety budget $d=100$, and deployment failure threshold $p^\star=0.05$.




\begin{table}[!htbp]
\centering
\caption{Reliability certificates for PPO on SafetyCar under bounded $L_\infty$ perturbations.}
\label{tab:car-linf-ppo-rel}
\small
\renewcommand{\arraystretch}{1.12}
\setlength{\tabcolsep}{5pt}

\begin{tabular*}{\textwidth}{@{\extracolsep{\fill}} l c r r r r c}
\toprule
\textbf{Policy}
& \textbf{$(k_a,k_s)$}
& \textbf{$\hat p_{\mathrm{task}}$}
& \textbf{$\hat p_{\mathrm{safety}}$}
& \textbf{$\hat p_F$}
& \textbf{B-CP UB}
& \textbf{Decision} \\
\midrule
PPO & $(0.00,0.00)$ & 0.0000 & 0.1952 & 0.1952 & 0.2044 & FAIL \\
PPO & $(0.00,0.03)$ & 0.0000 & 0.1948 & 0.1948 & 0.2040 & FAIL \\
PPO & $(0.00,0.06)$ & 0.0000 & 0.2017 & 0.2017 & 0.2110 & FAIL \\
PPO & $(0.03,0.00)$ & 0.0000 & 0.1969 & 0.1969 & 0.2062 & FAIL \\
PPO & $(0.03,0.03)$ & 0.0001 & 0.1994 & 0.1995 & 0.2088 & FAIL \\
PPO & $(0.03,0.06)$ & 0.0000 & 0.2024 & 0.2024 & 0.2118 & FAIL \\
PPO & $(0.06,0.00)$ & 0.0000 & 0.2006 & 0.2006 & 0.2099 & FAIL \\
PPO & $(0.06,0.03)$ & 0.0000 & 0.2007 & 0.2007 & 0.2100 & FAIL \\
PPO & $(0.06,0.06)$ & 0.0000 & 0.1915 & 0.1915 & 0.2007 & FAIL \\
\bottomrule
\end{tabular*}
\end{table}

\FloatBarrier




\begin{table}[!htbp]
\centering
\caption{Reliability certificates for PPO-Lagrangian on SafetyCar under bounded $L_\infty$ perturbations.}
\label{tab:car-linf-ppolag-rel}
\small
\renewcommand{\arraystretch}{1.12}
\setlength{\tabcolsep}{5pt}

\begin{tabular*}{\textwidth}{@{\extracolsep{\fill}} l c r r r r c}
\toprule
\textbf{Policy}
& \textbf{$(k_a,k_s)$}
& \textbf{$\hat p_{\mathrm{task}}$}
& \textbf{$\hat p_{\mathrm{safety}}$}
& \textbf{$\hat p_F$}
& \textbf{B-CP UB}
& \textbf{Decision} \\
\midrule
PPO-Lag & $(0.00,0.00)$ & 0.1010 & 0.0374 & 0.1369 & 0.1449 & FAIL \\
PPO-Lag & $(0.00,0.03)$ & 0.1048 & 0.0393 & 0.1425 & 0.1507 & FAIL \\
PPO-Lag & $(0.00,0.06)$ & 0.0993 & 0.0403 & 0.1380 & 0.1461 & FAIL \\
PPO-Lag & $(0.03,0.00)$ & 0.1036 & 0.0372 & 0.1391 & 0.1472 & FAIL \\
PPO-Lag & $(0.03,0.03)$ & 0.1037 & 0.0393 & 0.1411 & 0.1492 & FAIL \\
PPO-Lag & $(0.03,0.06)$ & 0.1009 & 0.0392 & 0.1377 & 0.1458 & FAIL \\
PPO-Lag & $(0.06,0.00)$ & 0.1039 & 0.0390 & 0.1418 & 0.1500 & FAIL \\
PPO-Lag & $(0.06,0.03)$ & 0.1053 & 0.0410 & 0.1442 & 0.1524 & FAIL \\
PPO-Lag & $(0.06,0.06)$ & 0.1001 & 0.0432 & 0.1419 & 0.1501 & FAIL \\
\bottomrule
\end{tabular*}
\end{table}

\FloatBarrier




\begin{table}[!htbp]
\centering
\caption{Reliability certificates for TRPO on SafetyCar under bounded $L_\infty$ perturbations.}
\label{tab:car-linf-trpo-rel}
\small
\renewcommand{\arraystretch}{1.12}
\setlength{\tabcolsep}{5pt}

\begin{tabular*}{\textwidth}{@{\extracolsep{\fill}} l c r r r r c}
\toprule
\textbf{Policy}
& \textbf{$(k_a,k_s)$}
& \textbf{$\hat p_{\mathrm{task}}$}
& \textbf{$\hat p_{\mathrm{safety}}$}
& \textbf{$\hat p_F$}
& \textbf{B-CP UB}
& \textbf{Decision} \\
\midrule
TRPO & $(0.00,0.00)$ & 0.0000 & 0.1970 & 0.1970 & 0.2063 & FAIL \\
TRPO & $(0.00,0.03)$ & 0.0000 & 0.1948 & 0.1948 & 0.2040 & FAIL \\
TRPO & $(0.00,0.06)$ & 0.0000 & 0.1971 & 0.1971 & 0.2064 & FAIL \\
TRPO & $(0.03,0.00)$ & 0.0000 & 0.1994 & 0.1994 & 0.2087 & FAIL \\
TRPO & $(0.03,0.03)$ & 0.0000 & 0.1949 & 0.1949 & 0.2041 & FAIL \\
TRPO & $(0.03,0.06)$ & 0.0000 & 0.1997 & 0.1997 & 0.2090 & FAIL \\
TRPO & $(0.06,0.00)$ & 0.0000 & 0.2003 & 0.2003 & 0.2096 & FAIL \\
TRPO & $(0.06,0.03)$ & 0.0000 & 0.1969 & 0.1969 & 0.2062 & FAIL \\
TRPO & $(0.06,0.06)$ & 0.0000 & 0.1912 & 0.1912 & 0.2004 & FAIL \\
\bottomrule
\end{tabular*}
\end{table}

\FloatBarrier


\begin{table}[!htbp]
\centering
\caption{Performance and cost certificates for TRPO-Lagrangian on SafetyCar under bounded $L_\infty$ perturbations.}
\label{tab:car-linf-trpolag-perf}
\small
\renewcommand{\arraystretch}{1.12}
\setlength{\tabcolsep}{5pt}

\begin{tabular*}{\textwidth}{@{\extracolsep{\fill}} l c r r r r}
\toprule
\textbf{Policy}
& \textbf{$(k_a,k_s)$}
& \textbf{$\bar{R}$}
& \textbf{$R_{\mathrm{H\text{-}LCB}}$}
& \textbf{$\bar{C}$}
& \textbf{$C_{\mathrm{H\text{-}UCB}}$} \\
\midrule
TRPO-Lag & $(0.00,0.00)$ & 23.35 & -276.64 & 28.88 & 43.88 \\
TRPO-Lag & $(0.00,0.03)$ & 23.57 & -276.42 & 29.42 & 44.42 \\
TRPO-Lag & $(0.00,0.06)$ & 23.47 & -276.52 & 29.70 & 44.70 \\
TRPO-Lag & $(0.03,0.00)$ & 23.38 & -276.61 & 29.93 & 44.93 \\
TRPO-Lag & $(0.03,0.03)$ & 23.33 & -276.66 & 29.46 & 44.46 \\
TRPO-Lag & $(0.03,0.06)$ & 23.56 & -276.44 & 29.37 & 44.37 \\
TRPO-Lag & $(0.06,0.00)$ & 23.44 & -276.55 & 28.95 & 43.95 \\
TRPO-Lag & $(0.06,0.03)$ & 23.48 & -276.51 & 29.76 & 44.76 \\
TRPO-Lag & $(0.06,0.06)$ & 23.45 & -276.54 & 29.29 & 44.29 \\
\bottomrule
\end{tabular*}
\end{table}

\begin{table}[!htbp]
\centering
\caption{Reliability certificates for TRPO-Lagrangian on SafetyCar under bounded $L_\infty$ perturbations.}
\label{tab:car-linf-trpolag-rel}
\small
\renewcommand{\arraystretch}{1.12}
\setlength{\tabcolsep}{5pt}

\begin{tabular*}{\textwidth}{@{\extracolsep{\fill}} l c r r r r c}
\toprule
\textbf{Policy}
& \textbf{$(k_a,k_s)$}
& \textbf{$\hat p_{\mathrm{task}}$}
& \textbf{$\hat p_{\mathrm{safety}}$}
& \textbf{$\hat p_F$}
& \textbf{B-CP UB}
& \textbf{Decision} \\
\midrule
TRPO-Lag & $(0.00,0.00)$ & 0.0164 & 0.0300 & 0.0460 & 0.0510 & FAIL \\
TRPO-Lag & $(0.00,0.03)$ & 0.0177 & 0.0272 & 0.0443 & \textbf{0.0492} & \textbf{PASS} \\
TRPO-Lag & $(0.00,0.06)$ & 0.0169 & 0.0311 & 0.0473 & 0.0524 & FAIL \\
TRPO-Lag & $(0.03,0.00)$ & 0.0167 & 0.0320 & 0.0480 & 0.0531 & FAIL \\
TRPO-Lag & $(0.03,0.03)$ & 0.0155 & 0.0330 & 0.0482 & 0.0533 & FAIL \\
TRPO-Lag & $(0.03,0.06)$ & 0.0158 & 0.0306 & 0.0458 & 0.0508 & FAIL \\
TRPO-Lag & $(0.06,0.00)$ & 0.0162 & 0.0289 & 0.0449 & \textbf{0.0499} & \textbf{PASS} \\
TRPO-Lag & $(0.06,0.03)$ & 0.0152 & 0.0309 & 0.0458 & 0.0508 & FAIL \\
TRPO-Lag & $(0.06,0.06)$ & 0.0160 & 0.0327 & 0.0482 & 0.0533 & FAIL \\
\bottomrule
\end{tabular*}
\end{table}

\FloatBarrier










\FloatBarrier

\subsection{Environment Perturbations}


\newcommand{\CarEnvPerfTable}[5]{%
\begin{table}[!htbp]
\centering
\caption{Performance and cost certificates for #1 on SafetyCar under #2.}
\label{#4}

\small
\renewcommand{\arraystretch}{1.12}
\setlength{\tabcolsep}{5pt}

\begin{tabular*}{\textwidth}{@{\extracolsep{\fill}} l c r r r r}
\toprule
\textbf{Policy}
& \textbf{#3}
& \textbf{$\bar{R}$}
& \textbf{$R_{\mathrm{H\text{-}LCB}}$}
& \textbf{$\bar{C}$}
& \textbf{$C_{\mathrm{H\text{-}UCB}}$} \\
\midrule
#5
\bottomrule
\end{tabular*}
\end{table}
}

\newcommand{\CarEnvRelTable}[5]{%
\begin{table}[!htbp]
\centering
\caption{Reliability certificates for #1 on SafetyCar under #2.}
\label{#4}

\small
\renewcommand{\arraystretch}{1.12}
\setlength{\tabcolsep}{5pt}

\begin{tabular*}{\textwidth}{@{\extracolsep{\fill}} l c r r r r c}
\toprule
\textbf{Policy}
& \textbf{#3}
& \textbf{$\hat p_{\mathrm{task}}$}
& \textbf{$\hat p_{\mathrm{safety}}$}
& \textbf{$\hat p_F$}
& \textbf{B-CP UB}
& \textbf{Decision} \\
\midrule
#5
\bottomrule
\end{tabular*}
\end{table}
}


\FloatBarrier
\subsubsection{Parameter Shift}

The parameter-shift experiments evaluate mass and friction scales
$m,f\in\{0.75,1.00,1.25\}$ with damping scale $\zeta=1.00$.
Each policy--configuration pair is evaluated using $N=10{,}000$
episodes. Certification uses a Bonferroni family size of $M=9$ per
policy, fixed safety budget $d=100$, and deployment failure threshold
$p^\star=0.05$.






\CarEnvRelTable{PPO}{parameter-shift perturbations}
{$(m,f,\zeta)$}
{tab:car-paramshift-ppo-rel}{%
PPO & $(0.75,0.75,1.00)$ & 0.0002 & 0.1932 & 0.1933 & 0.2025 & FAIL \\
PPO & $(0.75,1.00,1.00)$ & 0.0000 & 0.1999 & 0.1999 & 0.2092 & FAIL \\
PPO & $(0.75,1.25,1.00)$ & 0.0000 & 0.2122 & 0.2122 & 0.2217 & FAIL \\

\addlinespace[2pt]

PPO & $(1.00,0.75,1.00)$ & 0.0000 & 0.1830 & 0.1830 & 0.1920 & FAIL \\
PPO & $(1.00,1.00,1.00)$ & 0.0000 & 0.1992 & 0.1992 & 0.2085 & FAIL \\
PPO & $(1.00,1.25,1.00)$ & 0.0001 & 0.2163 & 0.2164 & 0.2260 & FAIL \\

\addlinespace[2pt]

PPO & $(1.25,0.75,1.00)$ & 0.0000 & 0.1819 & 0.1819 & 0.1909 & FAIL \\
PPO & $(1.25,1.00,1.00)$ & 0.0001 & 0.1966 & 0.1966 & 0.2059 & FAIL \\
PPO & $(1.25,1.25,1.00)$ & 0.0001 & 0.2164 & 0.2164 & 0.2260 & FAIL \\
}

\FloatBarrier






\CarEnvRelTable{PPO-Lagrangian}{parameter-shift perturbations}
{$(m,f,\zeta)$}
{tab:car-paramshift-ppolag-rel}{%
PPO-Lag & $(0.75,0.75,1.00)$ & 0.1427 & 0.0421 & 0.1814 & 0.1904 & FAIL \\
PPO-Lag & $(0.75,1.00,1.00)$ & 0.0987 & 0.0435 & 0.1411 & 0.1492 & FAIL \\
PPO-Lag & $(0.75,1.25,1.00)$ & 0.0671 & 0.0417 & 0.1082 & 0.1155 & FAIL \\

\addlinespace[2pt]

PPO-Lag & $(1.00,0.75,1.00)$ & 0.1448 & 0.0389 & 0.1804 & 0.1894 & FAIL \\
PPO-Lag & $(1.00,1.00,1.00)$ & 0.1024 & 0.0431 & 0.1446 & 0.1528 & FAIL \\
PPO-Lag & $(1.00,1.25,1.00)$ & 0.0693 & 0.0425 & 0.1109 & 0.1183 & FAIL \\

\addlinespace[2pt]

PPO-Lag & $(1.25,0.75,1.00)$ & 0.1470 & 0.0369 & 0.1809 & 0.1899 & FAIL \\
PPO-Lag & $(1.25,1.00,1.00)$ & 0.1045 & 0.0371 & 0.1403 & 0.1484 & FAIL \\
PPO-Lag & $(1.25,1.25,1.00)$ & 0.0748 & 0.0411 & 0.1151 & 0.1226 & FAIL \\
}

\FloatBarrier






\CarEnvRelTable{TRPO}{parameter-shift perturbations}
{$(m,f,\zeta)$}
{tab:car-paramshift-trpo-rel}{%
TRPO & $(0.75,0.75,1.00)$ & 0.0000 & 0.1966 & 0.1966 & 0.2059 & FAIL \\
TRPO & $(0.75,1.00,1.00)$ & 0.0000 & 0.2011 & 0.2011 & 0.2104 & FAIL \\
TRPO & $(0.75,1.25,1.00)$ & 0.0000 & 0.2191 & 0.2191 & 0.2287 & FAIL \\

\addlinespace[2pt]

TRPO & $(1.00,0.75,1.00)$ & 0.0000 & 0.1790 & 0.1790 & 0.1879 & FAIL \\
TRPO & $(1.00,1.00,1.00)$ & 0.0000 & 0.2007 & 0.2007 & 0.2100 & FAIL \\
TRPO & $(1.00,1.25,1.00)$ & 0.0000 & 0.2153 & 0.2153 & 0.2249 & FAIL \\

\addlinespace[2pt]

TRPO & $(1.25,0.75,1.00)$ & 0.0000 & 0.1744 & 0.1744 & 0.1833 & FAIL \\
TRPO & $(1.25,1.00,1.00)$ & 0.0000 & 0.2005 & 0.2005 & 0.2098 & FAIL \\
TRPO & $(1.25,1.25,1.00)$ & 0.0000 & 0.2176 & 0.2176 & 0.2272 & FAIL \\
}

\FloatBarrier






\CarEnvRelTable{TRPO-Lagrangian}{parameter-shift perturbations}
{$(m,f,\zeta)$}
{tab:car-paramshift-trpolag-rel}{%
TRPO-Lag & $(0.75,0.75,1.00)$ & 0.0179 & 0.0351 & 0.0525 & 0.0578 & FAIL \\
TRPO-Lag & $(0.75,1.00,1.00)$ & 0.0146 & 0.0339 & 0.0478 & 0.0529 & FAIL \\
TRPO-Lag & $(0.75,1.25,1.00)$ & 0.0140 & 0.0323 & 0.0459 & 0.0509 & FAIL \\

\addlinespace[2pt]

TRPO-Lag & $(1.00,0.75,1.00)$ & 0.0184 & 0.0289 & 0.0468 & 0.0519 & FAIL \\
TRPO-Lag & $(1.00,1.00,1.00)$ & 0.0168 & 0.0318 & 0.0481 & 0.0532 & FAIL \\
TRPO-Lag & $(1.00,1.25,1.00)$ & 0.0168 & 0.0314 & 0.0480 & 0.0531 & FAIL \\

\addlinespace[2pt]

TRPO-Lag & $(1.25,0.75,1.00)$ & 0.0196 & 0.0271 & 0.0461 & 0.0511 & FAIL \\
TRPO-Lag & $(1.25,1.00,1.00)$ & 0.0163 & 0.0307 & 0.0464 & 0.0514 & FAIL \\
TRPO-Lag & $(1.25,1.25,1.00)$ & 0.0164 & 0.0307 & 0.0467 & 0.0518 & FAIL \\
}

\FloatBarrier


\subsubsection{Randomized Dynamics}

The randomized-dynamics experiments evaluate
$r_{\mathrm{dyn}}\in\{0.00,0.10,0.25,0.50\}$ with $k_a=0$ and
$k_s\in\{0.00,0.05,0.10\}$. Each policy--configuration pair is
evaluated using $N=10{,}000$ episodes. Certification uses a Bonferroni
family size of $M=12$ per policy, fixed safety budget $d=100$, and
deployment failure threshold $p^\star=0.05$.








\CarEnvRelTable{PPO}{randomized-dynamics perturbations}
{$(r_{\mathrm{dyn}},k_a,k_s)$}
{tab:car-randdyn-ppo-rel}{%
PPO & $(0.00,0.00,0.00)$ & 0.0000 & 0.2013 & 0.2013 & 0.2111 & FAIL \\
PPO & $(0.00,0.00,0.05)$ & 0.5788 & 0.1536 & 0.6371 & 0.6486 & FAIL \\
PPO & $(0.00,0.00,0.10)$ & 0.7352 & 0.1432 & 0.7716 & 0.7816 & FAIL \\

\addlinespace[2pt]

PPO & $(0.10,0.00,0.00)$ & 0.0000 & 0.1973 & 0.1973 & 0.2070 & FAIL \\
PPO & $(0.10,0.00,0.05)$ & 0.5797 & 0.1541 & 0.6386 & 0.6501 & FAIL \\
PPO & $(0.10,0.00,0.10)$ & 0.7371 & 0.1465 & 0.7770 & 0.7869 & FAIL \\

\addlinespace[2pt]

PPO & $(0.25,0.00,0.00)$ & 0.0000 & 0.1952 & 0.1952 & 0.2049 & FAIL \\
PPO & $(0.25,0.00,0.05)$ & 0.5712 & 0.1535 & 0.6327 & 0.6442 & FAIL \\
PPO & $(0.25,0.00,0.10)$ & 0.7213 & 0.1461 & 0.7609 & 0.7711 & FAIL \\

\addlinespace[2pt]

PPO & $(0.50,0.00,0.00)$ & 0.0000 & 0.1978 & 0.1978 & 0.2075 & FAIL \\
PPO & $(0.50,0.00,0.05)$ & 0.5260 & 0.1458 & 0.5957 & 0.6075 & FAIL \\
PPO & $(0.50,0.00,0.10)$ & 0.6992 & 0.1451 & 0.7404 & 0.7508 & FAIL \\
}

\FloatBarrier








\CarEnvRelTable{PPO-Lagrangian}{randomized-dynamics perturbations}
{$(r_{\mathrm{dyn}},k_a,k_s)$}
{tab:car-randdyn-ppolag-rel}{%
PPO-Lag & $(0.00,0.00,0.00)$ & 0.1059 & 0.0414 & 0.1454 & 0.1540 & FAIL \\
PPO-Lag & $(0.00,0.00,0.05)$ & 0.7392 & 0.1447 & 0.7728 & 0.7828 & FAIL \\
PPO-Lag & $(0.00,0.00,0.10)$ & 0.8454 & 0.1891 & 0.8702 & 0.8781 & FAIL \\

\addlinespace[2pt]

PPO-Lag & $(0.10,0.00,0.00)$ & 0.1022 & 0.0400 & 0.1401 & 0.1486 & FAIL \\
PPO-Lag & $(0.10,0.00,0.05)$ & 0.7306 & 0.1487 & 0.7659 & 0.7760 & FAIL \\
PPO-Lag & $(0.10,0.00,0.10)$ & 0.8409 & 0.1843 & 0.8664 & 0.8744 & FAIL \\

\addlinespace[2pt]

PPO-Lag & $(0.25,0.00,0.00)$ & 0.1036 & 0.0369 & 0.1384 & 0.1469 & FAIL \\
PPO-Lag & $(0.25,0.00,0.05)$ & 0.7330 & 0.1380 & 0.7674 & 0.7775 & FAIL \\
PPO-Lag & $(0.25,0.00,0.10)$ & 0.8436 & 0.1728 & 0.8659 & 0.8740 & FAIL \\

\addlinespace[2pt]

PPO-Lag & $(0.50,0.00,0.00)$ & 0.1111 & 0.0382 & 0.1472 & 0.1559 & FAIL \\
PPO-Lag & $(0.50,0.00,0.05)$ & 0.7211 & 0.1199 & 0.7552 & 0.7654 & FAIL \\
PPO-Lag & $(0.50,0.00,0.10)$ & 0.8363 & 0.1319 & 0.8566 & 0.8649 & FAIL \\
}

\FloatBarrier








\CarEnvRelTable{TRPO}{randomized-dynamics perturbations}
{$(r_{\mathrm{dyn}},k_a,k_s)$}
{tab:car-randdyn-trpo-rel}{%
TRPO & $(0.00,0.00,0.00)$ & 0.0000 & 0.1952 & 0.1952 & 0.2049 & FAIL \\
TRPO & $(0.00,0.00,0.05)$ & 0.4956 & 0.1867 & 0.5906 & 0.6024 & FAIL \\
TRPO & $(0.00,0.00,0.10)$ & 0.7539 & 0.2162 & 0.7995 & 0.8090 & FAIL \\

\addlinespace[2pt]

TRPO & $(0.10,0.00,0.00)$ & 0.0000 & 0.1980 & 0.1980 & 0.2077 & FAIL \\
TRPO & $(0.10,0.00,0.05)$ & 0.4944 & 0.1812 & 0.5895 & 0.6013 & FAIL \\
TRPO & $(0.10,0.00,0.10)$ & 0.7476 & 0.2124 & 0.7943 & 0.8039 & FAIL \\

\addlinespace[2pt]

TRPO & $(0.25,0.00,0.00)$ & 0.0000 & 0.1933 & 0.1933 & 0.2029 & FAIL \\
TRPO & $(0.25,0.00,0.05)$ & 0.4952 & 0.1784 & 0.5825 & 0.5943 & FAIL \\
TRPO & $(0.25,0.00,0.10)$ & 0.7510 & 0.1900 & 0.7933 & 0.8029 & FAIL \\

\addlinespace[2pt]

TRPO & $(0.50,0.00,0.00)$ & 0.0000 & 0.1951 & 0.1951 & 0.2048 & FAIL \\
TRPO & $(0.50,0.00,0.05)$ & 0.4880 & 0.1670 & 0.5780 & 0.5898 & FAIL \\
TRPO & $(0.50,0.00,0.10)$ & 0.7412 & 0.1560 & 0.7804 & 0.7902 & FAIL \\
}

\FloatBarrier

\CarEnvPerfTable{TRPO-Lagrangian}{randomized-dynamics perturbations}
{$(r_{\mathrm{dyn}},k_a,k_s)$}
{tab:car-randdyn-trpolag-perf}{%
TRPO-Lag & $(0.00,0.00,0.00)$ & 23.58 & -285.85 & 28.92 & 44.39 \\
TRPO-Lag & $(0.00,0.00,0.05)$ & 0.53 & -308.90 & 32.90 & 48.37 \\
TRPO-Lag & $(0.00,0.00,0.10)$ & -1.07 & -310.50 & 37.52 & 53.00 \\

\addlinespace[2pt]

TRPO-Lag & $(0.10,0.00,0.00)$ & 23.31 & -286.12 & 29.67 & 45.14 \\
TRPO-Lag & $(0.10,0.00,0.05)$ & 0.55 & -308.88 & 33.41 & 48.88 \\
TRPO-Lag & $(0.10,0.00,0.10)$ & -1.04 & -310.48 & 36.65 & 52.12 \\

\addlinespace[2pt]

TRPO-Lag & $(0.25,0.00,0.00)$ & 22.81 & -286.63 & 28.86 & 44.33 \\
TRPO-Lag & $(0.25,0.00,0.05)$ & 0.62 & -308.81 & 34.01 & 49.48 \\
TRPO-Lag & $(0.25,0.00,0.10)$ & -0.95 & -310.39 & 37.21 & 52.68 \\

\addlinespace[2pt]

TRPO-Lag & $(0.50,0.00,0.00)$ & 20.92 & -288.51 & 30.40 & 45.87 \\
TRPO-Lag & $(0.50,0.00,0.05)$ & 0.76 & -308.68 & 34.04 & 49.51 \\
TRPO-Lag & $(0.50,0.00,0.10)$ & -0.77 & -310.21 & 36.63 & 52.10 \\
}

\CarEnvRelTable{TRPO-Lagrangian}{randomized-dynamics perturbations}
{$(r_{\mathrm{dyn}},k_a,k_s)$}
{tab:car-randdyn-trpolag-rel}{%
TRPO-Lag & $(0.00,0.00,0.00)$ & 0.0153 & 0.0310 & 0.0459 & 0.0512 & FAIL \\
TRPO-Lag & $(0.00,0.00,0.05)$ & 0.5041 & 0.1126 & 0.5648 & 0.5767 & FAIL \\
TRPO-Lag & $(0.00,0.00,0.10)$ & 0.7345 & 0.1399 & 0.7795 & 0.7894 & FAIL \\

\addlinespace[2pt]

TRPO-Lag & $(0.10,0.00,0.00)$ & 0.0177 & 0.0307 & 0.0477 & 0.0530 & FAIL \\
TRPO-Lag & $(0.10,0.00,0.05)$ & 0.5010 & 0.1120 & 0.5633 & 0.5752 & FAIL \\
TRPO-Lag & $(0.10,0.00,0.10)$ & 0.7241 & 0.1348 & 0.7664 & 0.7765 & FAIL \\

\addlinespace[2pt]

TRPO-Lag & $(0.25,0.00,0.00)$ & 0.0166 & 0.0273 & 0.0431 &
\textbf{0.0482} & \textbf{PASS} \\
TRPO-Lag & $(0.25,0.00,0.05)$ & 0.4834 & 0.1136 & 0.5476 & 0.5595 & FAIL \\
TRPO-Lag & $(0.25,0.00,0.10)$ & 0.7155 & 0.1396 & 0.7611 & 0.7712 & FAIL \\

\addlinespace[2pt]

TRPO-Lag & $(0.50,0.00,0.00)$ & 0.0178 & 0.0357 & 0.0525 & 0.0581 & FAIL \\
TRPO-Lag & $(0.50,0.00,0.05)$ & 0.4567 & 0.1115 & 0.5254 & 0.5374 & FAIL \\
TRPO-Lag & $(0.50,0.00,0.10)$ & 0.6980 & 0.1296 & 0.7431 & 0.7535 & FAIL \\
}

\FloatBarrier


\subsubsection{Delay/Latency}

The delay/latency experiments evaluate delays
$\Delta\in\{0,1,2,5\}$ timesteps with
$k_a\in\{0.00,0.05,0.10\}$ and $k_s=0$. Each policy--configuration
pair is evaluated using $N=10{,}000$ episodes. Certification uses a
Bonferroni family size of $M=12$ per policy, fixed safety budget $d=100$,
and deployment failure threshold $p^\star=0.05$.








\CarEnvRelTable{PPO}{delay/latency perturbations}
{$(\Delta,k_a,k_s)$}
{tab:car-delay-ppo-rel}{%
PPO & $(0,0.00,0.00)$ & 0.0000 & 0.1952 & 0.1952 & 0.2049 & FAIL \\
PPO & $(0,0.05,0.00)$ & 0.0001 & 0.1997 & 0.1998 & 0.2095 & FAIL \\
PPO & $(0,0.10,0.00)$ & 0.0000 & 0.1966 & 0.1966 & 0.2063 & FAIL \\

\addlinespace[2pt]

PPO & $(1,0.00,0.00)$ & 0.0000 & 0.1823 & 0.1823 & 0.1917 & FAIL \\
PPO & $(1,0.05,0.00)$ & 0.0000 & 0.1752 & 0.1752 & 0.1845 & FAIL \\
PPO & $(1,0.10,0.00)$ & 0.0000 & 0.1852 & 0.1852 & 0.1947 & FAIL \\

\addlinespace[2pt]

PPO & $(2,0.00,0.00)$ & 0.0000 & 0.1577 & 0.1577 & 0.1666 & FAIL \\
PPO & $(2,0.05,0.00)$ & 0.0000 & 0.1619 & 0.1619 & 0.1709 & FAIL \\
PPO & $(2,0.10,0.00)$ & 0.0000 & 0.1579 & 0.1579 & 0.1668 & FAIL \\

\addlinespace[2pt]

PPO & $(5,0.00,0.00)$ & 0.0000 & 0.1238 & 0.1238 & 0.1319 & FAIL \\
PPO & $(5,0.05,0.00)$ & 0.0000 & 0.1247 & 0.1247 & 0.1328 & FAIL \\
PPO & $(5,0.10,0.00)$ & 0.0000 & 0.1215 & 0.1215 & 0.1295 & FAIL \\
}

\FloatBarrier








\CarEnvRelTable{PPO-Lagrangian}{delay/latency perturbations}
{$(\Delta,k_a,k_s)$}
{tab:car-delay-ppolag-rel}{%
PPO-Lag & $(0,0.00,0.00)$ & 0.1010 & 0.0374 & 0.1369 & 0.1453 & FAIL \\
PPO-Lag & $(0,0.05,0.00)$ & 0.1046 & 0.0415 & 0.1445 & 0.1531 & FAIL \\
PPO-Lag & $(0,0.10,0.00)$ & 0.1025 & 0.0397 & 0.1409 & 0.1494 & FAIL \\

\addlinespace[2pt]

PPO-Lag & $(1,0.00,0.00)$ & 0.1428 & 0.0361 & 0.1761 & 0.1854 & FAIL \\
PPO-Lag & $(1,0.05,0.00)$ & 0.1394 & 0.0405 & 0.1775 & 0.1868 & FAIL \\
PPO-Lag & $(1,0.10,0.00)$ & 0.1416 & 0.0406 & 0.1794 & 0.1888 & FAIL \\

\addlinespace[2pt]

PPO-Lag & $(2,0.00,0.00)$ & 0.1195 & 0.0344 & 0.1517 & 0.1605 & FAIL \\
PPO-Lag & $(2,0.05,0.00)$ & 0.1215 & 0.0377 & 0.1567 & 0.1656 & FAIL \\
PPO-Lag & $(2,0.10,0.00)$ & 0.1198 & 0.0396 & 0.1565 & 0.1654 & FAIL \\

\addlinespace[2pt]

PPO-Lag & $(5,0.00,0.00)$ & 0.1269 & 0.0373 & 0.1625 & 0.1715 & FAIL \\
PPO-Lag & $(5,0.05,0.00)$ & 0.1287 & 0.0377 & 0.1640 & 0.1731 & FAIL \\
PPO-Lag & $(5,0.10,0.00)$ & 0.1281 & 0.0359 & 0.1624 & 0.1714 & FAIL \\
}









\CarEnvRelTable{TRPO}{delay/latency perturbations}
{$(\Delta,k_a,k_s)$}
{tab:car-delay-trpo-rel}{%
TRPO & $(0,0.00,0.00)$ & 0.0000 & 0.1970 & 0.1970 & 0.2067 & FAIL \\
TRPO & $(0,0.05,0.00)$ & 0.0000 & 0.1976 & 0.1976 & 0.2073 & FAIL \\
TRPO & $(0,0.10,0.00)$ & 0.0000 & 0.1968 & 0.1968 & 0.2065 & FAIL \\

\addlinespace[2pt]

TRPO & $(1,0.00,0.00)$ & 0.0000 & 0.1780 & 0.1780 & 0.1873 & FAIL \\
TRPO & $(1,0.05,0.00)$ & 0.0000 & 0.1773 & 0.1773 & 0.1866 & FAIL \\
TRPO & $(1,0.10,0.00)$ & 0.0000 & 0.1778 & 0.1778 & 0.1871 & FAIL \\

\addlinespace[2pt]

TRPO & $(2,0.00,0.00)$ & 0.0000 & 0.1535 & 0.1535 & 0.1623 & FAIL \\
TRPO & $(2,0.05,0.00)$ & 0.0000 & 0.1603 & 0.1603 & 0.1693 & FAIL \\
TRPO & $(2,0.10,0.00)$ & 0.0000 & 0.1613 & 0.1613 & 0.1703 & FAIL \\

\addlinespace[2pt]

TRPO & $(5,0.00,0.00)$ & 0.0000 & 0.1332 & 0.1332 & 0.1415 & FAIL \\
TRPO & $(5,0.05,0.00)$ & 0.0000 & 0.1315 & 0.1315 & 0.1398 & FAIL \\
TRPO & $(5,0.10,0.00)$ & 0.0000 & 0.1256 & 0.1256 & 0.1337 & FAIL \\
}

\FloatBarrier

\CarEnvPerfTable{TRPO-Lagrangian}{delay/latency perturbations}
{$(\Delta,k_a,k_s)$}
{tab:car-delay-trpolag-perf}{%
TRPO-Lag & $(0,0.00,0.00)$ & 23.35 & -286.08 & 28.88 & 44.35 \\
TRPO-Lag & $(0,0.05,0.00)$ & 23.30 & -286.13 & 29.60 & 45.07 \\
TRPO-Lag & $(0,0.10,0.00)$ & 23.53 & -285.91 & 28.99 & 44.46 \\

\addlinespace[2pt]

TRPO-Lag & $(1,0.00,0.00)$ & 22.98 & -286.46 & 28.14 & 43.61 \\
TRPO-Lag & $(1,0.05,0.00)$ & 23.28 & -286.15 & 29.10 & 44.57 \\
TRPO-Lag & $(1,0.10,0.00)$ & 23.43 & -286.01 & 29.11 & 44.59 \\

\addlinespace[2pt]

TRPO-Lag & $(2,0.00,0.00)$ & 21.90 & -287.54 & 27.87 & 43.34 \\
TRPO-Lag & $(2,0.05,0.00)$ & 22.20 & -287.24 & 28.45 & 43.92 \\
TRPO-Lag & $(2,0.10,0.00)$ & 22.29 & -287.14 & 28.10 & 43.57 \\

\addlinespace[2pt]

TRPO-Lag & $(5,0.00,0.00)$ & 18.59 & -290.85 & 28.51 & 43.98 \\
TRPO-Lag & $(5,0.05,0.00)$ & 18.88 & -290.55 & 28.63 & 44.10 \\
TRPO-Lag & $(5,0.10,0.00)$ & 19.10 & -290.33 & 28.28 & 43.76 \\
}

\CarEnvRelTable{TRPO-Lagrangian}{delay/latency perturbations}
{$(\Delta,k_a,k_s)$}
{tab:car-delay-trpolag-rel}{%
TRPO-Lag & $(0,0.00,0.00)$ & 0.0164 & 0.0300 & 0.0460 & 0.0513 & FAIL \\
TRPO-Lag & $(0,0.05,0.00)$ & 0.0182 & 0.0315 & 0.0490 & 0.0544 & FAIL \\
TRPO-Lag & $(0,0.10,0.00)$ & 0.0148 & 0.0311 & 0.0451 & 0.0503 & FAIL \\

\addlinespace[2pt]

TRPO-Lag & $(1,0.00,0.00)$ & 0.0171 & 0.0246 & 0.0412 &
\textbf{0.0462} & \textbf{PASS} \\
TRPO-Lag & $(1,0.05,0.00)$ & 0.0171 & 0.0294 & 0.0458 & 0.0510 & FAIL \\
TRPO-Lag & $(1,0.10,0.00)$ & 0.0165 & 0.0297 & 0.0455 & 0.0507 & FAIL \\

\addlinespace[2pt]

TRPO-Lag & $(2,0.00,0.00)$ & 0.0170 & 0.0251 & 0.0414 &
\textbf{0.0464} & \textbf{PASS} \\
TRPO-Lag & $(2,0.05,0.00)$ & 0.0156 & 0.0265 & 0.0419 &
\textbf{0.0469} & \textbf{PASS} \\
TRPO-Lag & $(2,0.10,0.00)$ & 0.0171 & 0.0257 & 0.0422 &
\textbf{0.0473} & \textbf{PASS} \\

\addlinespace[2pt]

TRPO-Lag & $(5,0.00,0.00)$ & 0.0163 & 0.0262 & 0.0420 &
\textbf{0.0470} & \textbf{PASS} \\
TRPO-Lag & $(5,0.05,0.00)$ & 0.0169 & 0.0250 & 0.0417 &
\textbf{0.0467} & \textbf{PASS} \\
TRPO-Lag & $(5,0.10,0.00)$ & 0.0167 & 0.0243 & 0.0404 &
\textbf{0.0454} & \textbf{PASS} \\
}

\FloatBarrier


\subsubsection{Layout Randomization}

The layout-randomization experiments evaluate
$\ell\in\{\text{nominal},\text{mild},\text{hard}\}$ with
$(k_a,k_s)\in
\{(0,0),(0.05,0),(0.10,0),(0,0.05),(0,0.10)\}$.
Each policy--configuration pair is evaluated using $N=10{,}000$
episodes. Certification uses a Bonferroni family size of $M=15$ per
policy, fixed safety budget $d=100$, and deployment failure threshold
$p^\star=0.05$.






\CarEnvRelTable{PPO}{layout-randomization perturbations}
{$(\ell,k_a,k_s)$}
{tab:car-layout-ppo-rel}{%
PPO & $(\text{nominal},0.00,0.00)$ & 0.0000 & 0.1952 & 0.1952 & 0.2052 & FAIL \\
PPO & $(\text{nominal},0.05,0.00)$ & 0.0001 & 0.1997 & 0.1998 & 0.2099 & FAIL \\
PPO & $(\text{nominal},0.10,0.00)$ & 0.0000 & 0.1966 & 0.1966 & 0.2066 & FAIL \\
PPO & $(\text{nominal},0.00,0.05)$ & 0.5885 & 0.1534 & 0.6481 & 0.6599 & FAIL \\
PPO & $(\text{nominal},0.00,0.10)$ & 0.7327 & 0.1455 & 0.7738 & 0.7841 & FAIL \\

\addlinespace[2pt]

PPO & $(\text{mild},0.00,0.00)$ & 0.0000 & 0.9094 & 0.9094 & 0.9164 & FAIL \\
PPO & $(\text{mild},0.05,0.00)$ & 0.0000 & 0.9048 & 0.9048 & 0.9119 & FAIL \\
PPO & $(\text{mild},0.10,0.00)$ & 0.0000 & 0.9143 & 0.9143 & 0.9211 & FAIL \\
PPO & $(\text{mild},0.00,0.05)$ & 0.6216 & 0.1408 & 0.6985 & 0.7098 & FAIL \\
PPO & $(\text{mild},0.00,0.10)$ & 0.7726 & 0.1575 & 0.8258 & 0.8351 & FAIL \\

\addlinespace[2pt]

PPO & $(\text{hard},0.00,0.00)$ & 0.0000 & 0.6826 & 0.6826 & 0.6941 & FAIL \\
PPO & $(\text{hard},0.05,0.00)$ & 0.0000 & 0.6728 & 0.6728 & 0.6844 & FAIL \\
PPO & $(\text{hard},0.10,0.00)$ & 0.0000 & 0.6838 & 0.6838 & 0.6953 & FAIL \\
PPO & $(\text{hard},0.00,0.05)$ & 0.6352 & 0.2071 & 0.7488 & 0.7595 & FAIL \\
PPO & $(\text{hard},0.00,0.10)$ & 0.7821 & 0.2480 & 0.8767 & 0.8847 & FAIL \\
}

\FloatBarrier






\CarEnvRelTable{PPO-Lagrangian}{layout-randomization perturbations}
{$(\ell,k_a,k_s)$}
{tab:car-layout-ppolag-rel}{%
PPO-Lag & $(\text{nominal},0.00,0.00)$ & 0.1010 & 0.0374 & 0.1369 & 0.1456 & FAIL \\
PPO-Lag & $(\text{nominal},0.05,0.00)$ & 0.1046 & 0.0415 & 0.1445 & 0.1534 & FAIL \\
PPO-Lag & $(\text{nominal},0.10,0.00)$ & 0.1025 & 0.0397 & 0.1409 & 0.1497 & FAIL \\
PPO-Lag & $(\text{nominal},0.00,0.05)$ & 0.7376 & 0.1469 & 0.7723 & 0.7826 & FAIL \\
PPO-Lag & $(\text{nominal},0.00,0.10)$ & 0.8406 & 0.1853 & 0.8654 & 0.8737 & FAIL \\

\addlinespace[2pt]

PPO-Lag & $(\text{mild},0.00,0.00)$ & 0.6155 & 0.0148 & 0.6287 & 0.6406 & FAIL \\
PPO-Lag & $(\text{mild},0.05,0.00)$ & 0.6179 & 0.0157 & 0.6322 & 0.6441 & FAIL \\
PPO-Lag & $(\text{mild},0.10,0.00)$ & 0.6158 & 0.0171 & 0.6313 & 0.6432 & FAIL \\
PPO-Lag & $(\text{mild},0.00,0.05)$ & 0.8004 & 0.1994 & 0.8735 & 0.8816 & FAIL \\
PPO-Lag & $(\text{mild},0.00,0.10)$ & 0.8871 & 0.1862 & 0.9257 & 0.9321 & FAIL \\

\addlinespace[2pt]

PPO-Lag & $(\text{hard},0.00,0.00)$ & 0.4565 & 0.0382 & 0.4931 & 0.5055 & FAIL \\
PPO-Lag & $(\text{hard},0.05,0.00)$ & 0.4540 & 0.0386 & 0.4907 & 0.5031 & FAIL \\
PPO-Lag & $(\text{hard},0.10,0.00)$ & 0.4547 & 0.0391 & 0.4919 & 0.5043 & FAIL \\
PPO-Lag & $(\text{hard},0.00,0.05)$ & 0.7972 & 0.2825 & 0.9125 & 0.9194 & FAIL \\
PPO-Lag & $(\text{hard},0.00,0.10)$ & 0.8933 & 0.2568 & 0.9575 & 0.9623 & FAIL \\
}

\FloatBarrier






\CarEnvRelTable{TRPO}{layout-randomization perturbations}
{$(\ell,k_a,k_s)$}
{tab:car-layout-trpo-rel}{%
TRPO & $(\text{nominal},0.00,0.00)$ & 0.0000 & 0.1970 & 0.1970 & 0.2070 & FAIL \\
TRPO & $(\text{nominal},0.05,0.00)$ & 0.0000 & 0.1976 & 0.1976 & 0.2076 & FAIL \\
TRPO & $(\text{nominal},0.10,0.00)$ & 0.0000 & 0.1968 & 0.1968 & 0.2068 & FAIL \\
TRPO & $(\text{nominal},0.00,0.05)$ & 0.4997 & 0.1943 & 0.5936 & 0.6058 & FAIL \\
TRPO & $(\text{nominal},0.00,0.10)$ & 0.7577 & 0.2233 & 0.8073 & 0.8170 & FAIL \\

\addlinespace[2pt]

TRPO & $(\text{mild},0.00,0.00)$ & 0.0000 & 0.9458 & 0.9458 & 0.9513 & FAIL \\
TRPO & $(\text{mild},0.05,0.00)$ & 0.0000 & 0.9429 & 0.9429 & 0.9485 & FAIL \\
TRPO & $(\text{mild},0.10,0.00)$ & 0.0000 & 0.9492 & 0.9492 & 0.9545 & FAIL \\
TRPO & $(\text{mild},0.00,0.05)$ & 0.5268 & 0.2206 & 0.6998 & 0.7111 & FAIL \\
TRPO & $(\text{mild},0.00,0.10)$ & 0.7752 & 0.2028 & 0.8803 & 0.8882 & FAIL \\

\addlinespace[2pt]

TRPO & $(\text{hard},0.00,0.00)$ & 0.0000 & 0.6739 & 0.6739 & 0.6855 & FAIL \\
TRPO & $(\text{hard},0.05,0.00)$ & 0.0000 & 0.6796 & 0.6796 & 0.6911 & FAIL \\
TRPO & $(\text{hard},0.10,0.00)$ & 0.0000 & 0.6736 & 0.6736 & 0.6852 & FAIL \\
TRPO & $(\text{hard},0.00,0.05)$ & 0.5618 & 0.3077 & 0.7866 & 0.7967 & FAIL \\
TRPO & $(\text{hard},0.00,0.10)$ & 0.7814 & 0.2653 & 0.9296 & 0.9358 & FAIL \\
}

\FloatBarrier






\CarEnvRelTable{TRPO-Lagrangian}{layout-randomization perturbations}
{$(\ell,k_a,k_s)$}
{tab:car-layout-trpolag-rel}{%
TRPO-Lag & $(\text{nominal},0.00,0.00)$ & 0.0164 & 0.0300 & 0.0460 & 0.0514 & FAIL \\
TRPO-Lag & $(\text{nominal},0.05,0.00)$ & 0.0182 & 0.0315 & 0.0490 & 0.0546 & FAIL \\
TRPO-Lag & $(\text{nominal},0.10,0.00)$ & 0.0148 & 0.0311 & 0.0451 & 0.0505 & FAIL \\
TRPO-Lag & $(\text{nominal},0.00,0.05)$ & 0.4989 & 0.1165 & 0.5620 & 0.5743 & FAIL \\
TRPO-Lag & $(\text{nominal},0.00,0.10)$ & 0.7289 & 0.1378 & 0.7705 & 0.7808 & FAIL \\

\addlinespace[2pt]

TRPO-Lag & $(\text{mild},0.00,0.00)$ & 0.1410 & 0.0481 & 0.1891 & 0.1990 & FAIL \\
TRPO-Lag & $(\text{mild},0.05,0.00)$ & 0.1347 & 0.0542 & 0.1889 & 0.1988 & FAIL \\
TRPO-Lag & $(\text{mild},0.10,0.00)$ & 0.1204 & 0.0539 & 0.1743 & 0.1839 & FAIL \\
TRPO-Lag & $(\text{mild},0.00,0.05)$ & 0.6053 & 0.1603 & 0.7002 & 0.7115 & FAIL \\
TRPO-Lag & $(\text{mild},0.00,0.10)$ & 0.7887 & 0.1755 & 0.8416 & 0.8505 & FAIL \\

\addlinespace[2pt]

TRPO-Lag & $(\text{hard},0.00,0.00)$ & 0.1110 & 0.0470 & 0.1580 & 0.1672 & FAIL \\
TRPO-Lag & $(\text{hard},0.05,0.00)$ & 0.1079 & 0.0449 & 0.1528 & 0.1619 & FAIL \\
TRPO-Lag & $(\text{hard},0.10,0.00)$ & 0.0968 & 0.0494 & 0.1462 & 0.1552 & FAIL \\
TRPO-Lag & $(\text{hard},0.00,0.05)$ & 0.5866 & 0.2041 & 0.7038 & 0.7151 & FAIL \\
TRPO-Lag & $(\text{hard},0.00,0.10)$ & 0.7850 & 0.2420 & 0.8649 & 0.8732 & FAIL \\
}

\FloatBarrier

\section{Complete Results for Safety Ant}
\label{App:Ant}


\FloatBarrier
\subsection{Gaussian Noise}

The Gaussian perturbation experiments evaluate state and action noise over
$k_a,k_s\in\{0.00,0.05,0.10\}$. Each policy--configuration pair is
evaluated using $N=10{,}000$ episodes. Certification uses a Bonferroni
family size of $M=9$ per policy, fixed safety budget $d=100$, and
deployment failure threshold $p^\star=0.05$.











\begin{table}[!htbp]
\centering
\caption{Reliability certificates for PPO on SafetyAnt under Gaussian perturbations.}
\label{tab:ant-gaussian-ppo-rel}

\small
\renewcommand{\arraystretch}{1.12}
\setlength{\tabcolsep}{5pt}

\begin{tabular*}{\textwidth}{@{\extracolsep{\fill}} l c r r r r c}
\toprule
\textbf{Policy}
& \textbf{$(k_a,k_s)$}
& \textbf{$\hat p_{\mathrm{task}}$}
& \textbf{$\hat p_{\mathrm{safety}}$}
& \textbf{$\hat p_F$}
& \textbf{B-CP UB}
& \textbf{Decision} \\
\midrule

PPO & $(0.00,0.00)$ & 0.1555 & 0.9751 & 0.9913 & 0.9933 & FAIL \\
PPO & $(0.00,0.05)$ & 0.8570 & 0.0000 & 0.8570 & 0.8649 & FAIL \\
PPO & $(0.00,0.10)$ & 0.8603 & 0.0000 & 0.8603 & 0.8681 & FAIL \\

\addlinespace[2pt]

PPO & $(0.05,0.00)$ & 0.2569 & 0.9556 & 0.9875 & 0.9899 & FAIL \\
PPO & $(0.05,0.05)$ & 0.8575 & 0.0000 & 0.8575 & 0.8654 & FAIL \\
PPO & $(0.05,0.10)$ & 0.8716 & 0.0000 & 0.8716 & 0.8792 & FAIL \\

\addlinespace[2pt]

PPO & $(0.10,0.00)$ & 0.5951 & 0.8455 & 0.9712 & 0.9749 & FAIL \\
PPO & $(0.10,0.05)$ & 0.8752 & 0.0000 & 0.8752 & 0.8827 & FAIL \\
PPO & $(0.10,0.10)$ & 0.8779 & 0.0000 & 0.8779 & 0.8853 & FAIL \\

\bottomrule
\end{tabular*}
\end{table}

\FloatBarrier











\begin{table}[!htbp]
\centering
\caption{Reliability certificates for PPO-Lagrangian on SafetyAnt under Gaussian perturbations.}
\label{tab:ant-gaussian-ppolag-rel}

\small
\renewcommand{\arraystretch}{1.12}
\setlength{\tabcolsep}{5pt}

\begin{tabular*}{\textwidth}{@{\extracolsep{\fill}} l c r r r r c}
\toprule
\textbf{Policy}
& \textbf{$(k_a,k_s)$}
& \textbf{$\hat p_{\mathrm{task}}$}
& \textbf{$\hat p_{\mathrm{safety}}$}
& \textbf{$\hat p_F$}
& \textbf{B-CP UB}
& \textbf{Decision} \\
\midrule

PPO-Lag & $(0.00,0.00)$ & 0.0158 & 0.7972 & 0.8093 & 0.8182 & FAIL \\
PPO-Lag & $(0.00,0.05)$ & 0.6255 & 0.0000 & 0.6255 & 0.6366 & FAIL \\
PPO-Lag & $(0.00,0.10)$ & 0.6412 & 0.0000 & 0.6412 & 0.6522 & FAIL \\

\addlinespace[2pt]

PPO-Lag & $(0.05,0.00)$ & 0.2548 & 0.4744 & 0.6802 & 0.6908 & FAIL \\
PPO-Lag & $(0.05,0.05)$ & 0.6905 & 0.0000 & 0.6905 & 0.7010 & FAIL \\
PPO-Lag & $(0.05,0.10)$ & 0.6991 & 0.0000 & 0.6991 & 0.7096 & FAIL \\

\addlinespace[2pt]

PPO-Lag & $(0.10,0.00)$ & 0.7672 & 0.0005 & 0.7676 & 0.7772 & FAIL \\
PPO-Lag & $(0.10,0.05)$ & 0.7847 & 0.0000 & 0.7847 & 0.7940 & FAIL \\
PPO-Lag & $(0.10,0.10)$ & 0.7965 & 0.0000 & 0.7965 & 0.8056 & FAIL \\

\bottomrule
\end{tabular*}
\end{table}

\FloatBarrier











\begin{table}[!htbp]
\centering
\caption{Reliability certificates for TRPO on SafetyAnt under Gaussian perturbations.}
\label{tab:ant-gaussian-trpo-rel}

\small
\renewcommand{\arraystretch}{1.12}
\setlength{\tabcolsep}{5pt}

\begin{tabular*}{\textwidth}{@{\extracolsep{\fill}} l c r r r r c}
\toprule
\textbf{Policy}
& \textbf{$(k_a,k_s)$}
& \textbf{$\hat p_{\mathrm{task}}$}
& \textbf{$\hat p_{\mathrm{safety}}$}
& \textbf{$\hat p_F$}
& \textbf{B-CP UB}
& \textbf{Decision} \\
\midrule

TRPO & $(0.00,0.00)$ & 0.0146 & 0.9978 & 0.9988 & 0.9994 & FAIL \\
TRPO & $(0.00,0.05)$ & 0.9356 & 0.0000 & 0.9356 & 0.9411 & FAIL \\
TRPO & $(0.00,0.10)$ & 0.9408 & 0.0000 & 0.9408 & 0.9461 & FAIL \\

\addlinespace[2pt]

TRPO & $(0.05,0.00)$ & 0.0600 & 0.9931 & 0.9983 & 0.9991 & FAIL \\
TRPO & $(0.05,0.05)$ & 0.9351 & 0.0000 & 0.9351 & 0.9406 & FAIL \\
TRPO & $(0.05,0.10)$ & 0.9417 & 0.0000 & 0.9417 & 0.9469 & FAIL \\

\addlinespace[2pt]

TRPO & $(0.10,0.00)$ & 0.3573 & 0.9273 & 0.9866 & 0.9891 & FAIL \\
TRPO & $(0.10,0.05)$ & 0.9374 & 0.0000 & 0.9374 & 0.9428 & FAIL \\
TRPO & $(0.10,0.10)$ & 0.9394 & 0.0000 & 0.9394 & 0.9447 & FAIL \\

\bottomrule
\end{tabular*}
\end{table}

\FloatBarrier


\begin{table}[!htbp]
\centering
\caption{Performance and cost certificates for TRPO-Lagrangian on SafetyAnt under Gaussian perturbations.}
\label{tab:ant-gaussian-trpolag-perf}

\small
\renewcommand{\arraystretch}{1.12}
\setlength{\tabcolsep}{5pt}

\begin{tabular*}{\textwidth}{@{\extracolsep{\fill}} l c r r r r}
\toprule
\textbf{Policy}
& \textbf{$(k_a,k_s)$}
& \textbf{$\bar{R}$}
& \textbf{$R_{\mathrm{H\text{-}LCB}}$}
& \textbf{$\bar{C}$}
& \textbf{$C_{\mathrm{H\text{-}UCB}}$} \\
\midrule

TRPO-Lag & $(0.00,0.00)$ & 2863.30 & 2563.30 & 29.67 & 44.67 \\
TRPO-Lag & $(0.00,0.05)$ & -51.65 & -351.65 & 2.55 & 17.55 \\
TRPO-Lag & $(0.00,0.10)$ & -56.38 & -356.37 & 2.42 & 17.42 \\

\addlinespace[2pt]

TRPO-Lag & $(0.05,0.00)$ & 2069.27 & 1769.27 & 31.17 & 46.17 \\
TRPO-Lag & $(0.05,0.05)$ & -56.09 & -356.08 & 2.45 & 17.45 \\
TRPO-Lag & $(0.05,0.10)$ & -56.72 & -356.72 & 2.18 & 17.18 \\

\addlinespace[2pt]

TRPO-Lag & $(0.10,0.00)$ & 1198.36 & 898.37 & 18.55 & 33.55 \\
TRPO-Lag & $(0.10,0.05)$ & -63.90 & -363.89 & 2.20 & 17.20 \\
TRPO-Lag & $(0.10,0.10)$ & -65.74 & -365.74 & 2.03 & 17.03 \\

\bottomrule
\end{tabular*}
\end{table}

\begin{table}[!htbp]
\centering
\caption{Reliability certificates for TRPO-Lagrangian on SafetyAnt under Gaussian perturbations.}
\label{tab:ant-gaussian-trpolag-rel}

\small
\renewcommand{\arraystretch}{1.12}
\setlength{\tabcolsep}{5pt}

\begin{tabular*}{\textwidth}{@{\extracolsep{\fill}} l c r r r r c}
\toprule
\textbf{Policy}
& \textbf{$(k_a,k_s)$}
& \textbf{$\hat p_{\mathrm{task}}$}
& \textbf{$\hat p_{\mathrm{safety}}$}
& \textbf{$\hat p_F$}
& \textbf{B-CP UB}
& \textbf{Decision} \\
\midrule

TRPO-Lag & $(0.00,0.00)$ & 0.0081 & 0.0000 & 0.0081 & \textbf{0.0104} & \textbf{PASS} \\
TRPO-Lag & $(0.00,0.05)$ & 0.9198 & 0.0000 & 0.9198 & 0.9259 & FAIL \\
TRPO-Lag & $(0.00,0.10)$ & 0.9200 & 0.0000 & 0.9200 & 0.9261 & FAIL \\

\addlinespace[2pt]

TRPO-Lag & $(0.05,0.00)$ & 0.1304 & 0.0008 & 0.1312 & 0.1391 & FAIL \\
TRPO-Lag & $(0.05,0.05)$ & 0.9188 & 0.0000 & 0.9188 & 0.9249 & FAIL \\
TRPO-Lag & $(0.05,0.10)$ & 0.9274 & 0.0000 & 0.9274 & 0.9332 & FAIL \\

\addlinespace[2pt]

TRPO-Lag & $(0.10,0.00)$ & 0.3600 & 0.0000 & 0.3600 & 0.3711 & FAIL \\
TRPO-Lag & $(0.10,0.05)$ & 0.9214 & 0.0000 & 0.9214 & 0.9274 & FAIL \\
TRPO-Lag & $(0.10,0.10)$ & 0.9252 & 0.0000 & 0.9252 & 0.9311 & FAIL \\

\bottomrule
\end{tabular*}
\end{table}

\FloatBarrier





\subsection{Uniform Noise}

The Uniform perturbation experiments evaluate state and action noise over
$k_a,k_s\in\{0.00,0.05,0.10\}$. Each policy--configuration pair is
evaluated using $N=10{,}000$ episodes. Certification uses a Bonferroni
family size of $M=9$ per policy, fixed safety budget $d=100$, and
deployment failure threshold $p^\star=0.05$.










\begin{table}[!htbp]
\centering
\caption{Reliability certificates for PPO on SafetyAnt under Uniform perturbations.}
\label{tab:ant-uniform-ppo-rel}

\small
\renewcommand{\arraystretch}{1.12}
\setlength{\tabcolsep}{5pt}

\begin{tabular*}{\textwidth}{@{\extracolsep{\fill}} l c r r r r c}
\toprule
\textbf{Policy}
& \textbf{$(k_a,k_s)$}
& \textbf{$\hat p_{\mathrm{task}}$}
& \textbf{$\hat p_{\mathrm{safety}}$}
& \textbf{$\hat p_F$}
& \textbf{B-CP UB}
& \textbf{Decision} \\
\midrule

PPO & $(0.00,0.00)$ & 0.1555 & 0.9751 & 0.9913 & 0.9933 & FAIL \\
PPO & $(0.00,0.05)$ & 0.8496 & 0.0000 & 0.8496 & 0.8577 & FAIL \\
PPO & $(0.00,0.10)$ & 0.8567 & 0.0000 & 0.8567 & 0.8646 & FAIL \\

\addlinespace[2pt]

PPO & $(0.05,0.00)$ & 0.1831 & 0.9678 & 0.9889 & 0.9912 & FAIL \\
PPO & $(0.05,0.05)$ & 0.8469 & 0.0000 & 0.8469 & 0.8550 & FAIL \\
PPO & $(0.05,0.10)$ & 0.8560 & 0.0000 & 0.8560 & 0.8639 & FAIL \\

\addlinespace[2pt]

PPO & $(0.10,0.00)$ & 0.2888 & 0.9482 & 0.9850 & 0.9876 & FAIL \\
PPO & $(0.10,0.05)$ & 0.8536 & 0.0000 & 0.8536 & 0.8616 & FAIL \\
PPO & $(0.10,0.10)$ & 0.8680 & 0.0000 & 0.8680 & 0.8756 & FAIL \\

\bottomrule
\end{tabular*}
\end{table}

\FloatBarrier










\begin{table}[!htbp]
\centering
\caption{Reliability certificates for PPO-Lagrangian on SafetyAnt under Uniform perturbations.}
\label{tab:ant-uniform-ppolag-rel}

\small
\renewcommand{\arraystretch}{1.12}
\setlength{\tabcolsep}{5pt}

\begin{tabular*}{\textwidth}{@{\extracolsep{\fill}} l c r r r r c}
\toprule
\textbf{Policy}
& \textbf{$(k_a,k_s)$}
& \textbf{$\hat p_{\mathrm{task}}$}
& \textbf{$\hat p_{\mathrm{safety}}$}
& \textbf{$\hat p_F$}
& \textbf{B-CP UB}
& \textbf{Decision} \\
\midrule

PPO-Lag & $(0.00,0.00)$ & 0.0158 & 0.7972 & 0.8093 & 0.8182 & FAIL \\
PPO-Lag & $(0.00,0.05)$ & 0.5858 & 0.0000 & 0.5858 & 0.5971 & FAIL \\
PPO-Lag & $(0.00,0.10)$ & 0.6342 & 0.0000 & 0.6342 & 0.6452 & FAIL \\

\addlinespace[2pt]

PPO-Lag & $(0.05,0.00)$ & 0.0354 & 0.7613 & 0.7847 & 0.7940 & FAIL \\
PPO-Lag & $(0.05,0.05)$ & 0.6110 & 0.0000 & 0.6110 & 0.6222 & FAIL \\
PPO-Lag & $(0.05,0.10)$ & 0.6546 & 0.0000 & 0.6546 & 0.6655 & FAIL \\

\addlinespace[2pt]

PPO-Lag & $(0.10,0.00)$ & 0.3803 & 0.2974 & 0.6432 & 0.6541 & FAIL \\
PPO-Lag & $(0.10,0.05)$ & 0.6709 & 0.0000 & 0.6709 & 0.6816 & FAIL \\
PPO-Lag & $(0.10,0.10)$ & 0.6929 & 0.0000 & 0.6929 & 0.7034 & FAIL \\

\bottomrule
\end{tabular*}
\end{table}

\FloatBarrier










\begin{table}[!htbp]
\centering
\caption{Reliability certificates for TRPO on SafetyAnt under Uniform perturbations.}
\label{tab:ant-uniform-trpo-rel}

\small
\renewcommand{\arraystretch}{1.12}
\setlength{\tabcolsep}{5pt}

\begin{tabular*}{\textwidth}{@{\extracolsep{\fill}} l c r r r r c}
\toprule
\textbf{Policy}
& \textbf{$(k_a,k_s)$}
& \textbf{$\hat p_{\mathrm{task}}$}
& \textbf{$\hat p_{\mathrm{safety}}$}
& \textbf{$\hat p_F$}
& \textbf{B-CP UB}
& \textbf{Decision} \\
\midrule

TRPO & $(0.00,0.00)$ & 0.0146 & 0.9978 & 0.9988 & 0.9994 & FAIL \\
TRPO & $(0.00,0.05)$ & 0.9272 & 0.0000 & 0.9272 & 0.9330 & FAIL \\
TRPO & $(0.00,0.10)$ & 0.9371 & 0.0000 & 0.9371 & 0.9425 & FAIL \\

\addlinespace[2pt]

TRPO & $(0.05,0.00)$ & 0.0267 & 0.9952 & 0.9981 & 0.9990 & FAIL \\
TRPO & $(0.05,0.05)$ & 0.9254 & 0.0000 & 0.9254 & 0.9313 & FAIL \\
TRPO & $(0.05,0.10)$ & 0.9322 & 0.0000 & 0.9322 & 0.9378 & FAIL \\

\addlinespace[2pt]

TRPO & $(0.10,0.00)$ & 0.0890 & 0.9884 & 0.9970 & 0.9981 & FAIL \\
TRPO & $(0.10,0.05)$ & 0.9206 & 0.0000 & 0.9206 & 0.9267 & FAIL \\
TRPO & $(0.10,0.10)$ & 0.9377 & 0.0000 & 0.9377 & 0.9431 & FAIL \\

\bottomrule
\end{tabular*}
\end{table}

\FloatBarrier

\begin{table}[!htbp]
\centering
\caption{Performance and cost certificates for TRPO-Lagrangian on SafetyAnt under Uniform perturbations.}
\label{tab:ant-uniform-trpolag-perf}

\small
\renewcommand{\arraystretch}{1.12}
\setlength{\tabcolsep}{5pt}

\begin{tabular*}{\textwidth}{@{\extracolsep{\fill}} l c r r r r}
\toprule
\textbf{Policy}
& \textbf{$(k_a,k_s)$}
& \textbf{$\bar{R}$}
& \textbf{$R_{\mathrm{H\text{-}LCB}}$}
& \textbf{$\bar{C}$}
& \textbf{$C_{\mathrm{H\text{-}UCB}}$} \\
\midrule

TRPO-Lag & $(0.00,0.00)$ & 2863.30 & 2563.30 & 29.67 & 44.67 \\
TRPO-Lag & $(0.00,0.05)$ & -45.05 & -345.04 & 3.08 & 18.08 \\
TRPO-Lag & $(0.00,0.10)$ & -52.60 & -352.59 & 2.49 & 17.49 \\

\addlinespace[2pt]

TRPO-Lag & $(0.05,0.00)$ & 2559.84 & 2259.84 & 43.21 & 58.21 \\
TRPO-Lag & $(0.05,0.05)$ & -46.45 & -346.45 & 3.07 & 18.07 \\
TRPO-Lag & $(0.05,0.10)$ & -54.28 & -354.27 & 2.49 & 17.49 \\

\addlinespace[2pt]

TRPO-Lag & $(0.10,0.00)$ & 1955.85 & 1655.85 & 28.30 & 43.30 \\
TRPO-Lag & $(0.10,0.05)$ & -50.11 & -350.10 & 2.88 & 17.88 \\
TRPO-Lag & $(0.10,0.10)$ & -56.65 & -356.65 & 2.38 & 17.38 \\

\bottomrule
\end{tabular*}
\end{table}

\begin{table}[!htbp]
\centering
\caption{Reliability certificates for TRPO-Lagrangian on SafetyAnt under Uniform perturbations.}
\label{tab:ant-uniform-trpolag-rel}

\small
\renewcommand{\arraystretch}{1.12}
\setlength{\tabcolsep}{5pt}

\begin{tabular*}{\textwidth}{@{\extracolsep{\fill}} l c r r r r c}
\toprule
\textbf{Policy}
& \textbf{$(k_a,k_s)$}
& \textbf{$\hat p_{\mathrm{task}}$}
& \textbf{$\hat p_{\mathrm{safety}}$}
& \textbf{$\hat p_F$}
& \textbf{B-CP UB}
& \textbf{Decision} \\
\midrule

TRPO-Lag & $(0.00,0.00)$ & 0.0081 & 0.0000 & 0.0081 & \textbf{0.0104} & \textbf{PASS} \\
TRPO-Lag & $(0.00,0.05)$ & 0.9124 & 0.0000 & 0.9124 & 0.9188 & FAIL \\
TRPO-Lag & $(0.00,0.10)$ & 0.9231 & 0.0000 & 0.9231 & 0.9291 & FAIL \\

\addlinespace[2pt]

TRPO-Lag & $(0.05,0.00)$ & 0.0334 & 0.0008 & 0.0341 & \textbf{0.0385} & \textbf{PASS} \\
TRPO-Lag & $(0.05,0.05)$ & 0.9094 & 0.0000 & 0.9094 & 0.9159 & FAIL \\
TRPO-Lag & $(0.05,0.10)$ & 0.9210 & 0.0000 & 0.9210 & 0.9271 & FAIL \\

\addlinespace[2pt]

TRPO-Lag & $(0.10,0.00)$ & 0.1621 & 0.0003 & 0.1623 & 0.1709 & FAIL \\
TRPO-Lag & $(0.10,0.05)$ & 0.9158 & 0.0000 & 0.9158 & 0.9220 & FAIL \\
TRPO-Lag & $(0.10,0.10)$ & 0.9247 & 0.0000 & 0.9247 & 0.9306 & FAIL \\

\bottomrule
\end{tabular*}
\end{table}

\FloatBarrier





\FloatBarrier

\subsection{Bounded Adversarial $L_\infty$ Noise}

The bounded $L_\infty$ perturbation experiments evaluate state and action
perturbations over $k_a,k_s\in\{0.00,0.03,0.06\}$. Each
policy--configuration pair is evaluated using $N=10{,}000$ episodes.
Certification uses a Bonferroni family size of $M=9$ per policy, fixed
safety budget $d=100$, and deployment failure threshold $p^\star=0.05$.











\begin{table}[!htbp]
\centering
\caption{Reliability certificates for PPO on SafetyAnt under bounded $L_\infty$ perturbations.}
\label{tab:ant-linf-ppo-rel}

\small
\renewcommand{\arraystretch}{1.12}
\setlength{\tabcolsep}{5pt}

\begin{tabular*}{\textwidth}{@{\extracolsep{\fill}} l c r r r r c}
\toprule
\textbf{Policy}
& \textbf{$(k_a,k_s)$}
& \textbf{$\hat p_{\mathrm{task}}$}
& \textbf{$\hat p_{\mathrm{safety}}$}
& \textbf{$\hat p_F$}
& \textbf{B-CP UB}
& \textbf{Decision} \\
\midrule

PPO & $(0.00,0.00)$ & 0.1466 & 0.9770 & 0.9922 & 0.9941 & FAIL \\
PPO & $(0.00,0.03)$ & 0.1509 & 0.9760 & 0.9929 & 0.9947 & FAIL \\
PPO & $(0.00,0.06)$ & 0.1703 & 0.9717 & 0.9916 & 0.9935 & FAIL \\

\addlinespace[2pt]

PPO & $(0.03,0.00)$ & 0.1537 & 0.9723 & 0.9905 & 0.9926 & FAIL \\
PPO & $(0.03,0.03)$ & 0.1568 & 0.9737 & 0.9910 & 0.9930 & FAIL \\
PPO & $(0.03,0.06)$ & 0.1655 & 0.9709 & 0.9909 & 0.9929 & FAIL \\

\addlinespace[2pt]

PPO & $(0.06,0.00)$ & 0.1809 & 0.9701 & 0.9900 & 0.9921 & FAIL \\
PPO & $(0.06,0.03)$ & 0.1788 & 0.9697 & 0.9892 & 0.9914 & FAIL \\
PPO & $(0.06,0.06)$ & 0.1898 & 0.9705 & 0.9906 & 0.9927 & FAIL \\

\bottomrule
\end{tabular*}
\end{table}

\FloatBarrier











\begin{table}[!htbp]
\centering
\caption{Reliability certificates for PPO-Lagrangian on SafetyAnt under bounded $L_\infty$ perturbations.}
\label{tab:ant-linf-ppolag-rel}

\small
\renewcommand{\arraystretch}{1.12}
\setlength{\tabcolsep}{5pt}

\begin{tabular*}{\textwidth}{@{\extracolsep{\fill}} l c r r r r c}
\toprule
\textbf{Policy}
& \textbf{$(k_a,k_s)$}
& \textbf{$\hat p_{\mathrm{task}}$}
& \textbf{$\hat p_{\mathrm{safety}}$}
& \textbf{$\hat p_F$}
& \textbf{B-CP UB}
& \textbf{Decision} \\
\midrule

PPO-Lag & $(0.00,0.00)$ & 0.0169 & 0.7966 & 0.8088 & 0.8177 & FAIL \\
PPO-Lag & $(0.00,0.03)$ & 0.0167 & 0.8003 & 0.8108 & 0.8197 & FAIL \\
PPO-Lag & $(0.00,0.06)$ & 0.0221 & 0.7818 & 0.7962 & 0.8053 & FAIL \\

\addlinespace[2pt]

PPO-Lag & $(0.03,0.00)$ & 0.0150 & 0.7909 & 0.8010 & 0.8101 & FAIL \\
PPO-Lag & $(0.03,0.03)$ & 0.0207 & 0.7864 & 0.8008 & 0.8099 & FAIL \\
PPO-Lag & $(0.03,0.06)$ & 0.0269 & 0.7689 & 0.7855 & 0.7948 & FAIL \\

\addlinespace[2pt]

PPO-Lag & $(0.06,0.00)$ & 0.0357 & 0.7467 & 0.7698 & 0.7794 & FAIL \\
PPO-Lag & $(0.06,0.03)$ & 0.0434 & 0.7376 & 0.7659 & 0.7755 & FAIL \\
PPO-Lag & $(0.06,0.06)$ & 0.0578 & 0.7056 & 0.7448 & 0.7547 & FAIL \\

\bottomrule
\end{tabular*}
\end{table}

\FloatBarrier











\begin{table}[!htbp]
\centering
\caption{Reliability certificates for TRPO on SafetyAnt under bounded $L_\infty$ perturbations.}
\label{tab:ant-linf-trpo-rel}

\small
\renewcommand{\arraystretch}{1.12}
\setlength{\tabcolsep}{5pt}

\begin{tabular*}{\textwidth}{@{\extracolsep{\fill}} l c r r r r c}
\toprule
\textbf{Policy}
& \textbf{$(k_a,k_s)$}
& \textbf{$\hat p_{\mathrm{task}}$}
& \textbf{$\hat p_{\mathrm{safety}}$}
& \textbf{$\hat p_F$}
& \textbf{B-CP UB}
& \textbf{Decision} \\
\midrule

TRPO & $(0.00,0.00)$ & 0.0185 & 0.9979 & 0.9992 & 0.9997 & FAIL \\
TRPO & $(0.00,0.03)$ & 0.0160 & 0.9970 & 0.9988 & 0.9994 & FAIL \\
TRPO & $(0.00,0.06)$ & 0.0231 & 0.9969 & 0.9991 & 0.9996 & FAIL \\

\addlinespace[2pt]

TRPO & $(0.03,0.00)$ & 0.0190 & 0.9977 & 0.9990 & 0.9996 & FAIL \\
TRPO & $(0.03,0.03)$ & 0.0207 & 0.9973 & 0.9990 & 0.9996 & FAIL \\
TRPO & $(0.03,0.06)$ & 0.0269 & 0.9959 & 0.9987 & 0.9994 & FAIL \\

\addlinespace[2pt]

TRPO & $(0.06,0.00)$ & 0.0257 & 0.9958 & 0.9988 & 0.9994 & FAIL \\
TRPO & $(0.06,0.03)$ & 0.0260 & 0.9963 & 0.9991 & 0.9996 & FAIL \\
TRPO & $(0.06,0.06)$ & 0.0329 & 0.9945 & 0.9979 & 0.9988 & FAIL \\

\bottomrule
\end{tabular*}
\end{table}

\FloatBarrier


\begin{table}[!htbp]
\centering
\caption{Performance and cost certificates for TRPO-Lagrangian on SafetyAnt under bounded $L_\infty$ perturbations.}
\label{tab:ant-linf-trpolag-perf}

\small
\renewcommand{\arraystretch}{1.12}
\setlength{\tabcolsep}{5pt}

\begin{tabular*}{\textwidth}{@{\extracolsep{\fill}} l c r r r r}
\toprule
\textbf{Policy}
& \textbf{$(k_a,k_s)$}
& \textbf{$\bar{R}$}
& \textbf{$R_{\mathrm{H\text{-}LCB}}$}
& \textbf{$\bar{C}$}
& \textbf{$C_{\mathrm{H\text{-}UCB}}$} \\
\midrule

TRPO-Lag & $(0.00,0.00)$ & 2880.78 & 2580.78 & 29.78 & 44.78 \\
TRPO-Lag & $(0.00,0.03)$ & 2789.97 & 2489.98 & 37.15 & 52.15 \\
TRPO-Lag & $(0.00,0.06)$ & 2539.32 & 2239.33 & 48.75 & 63.75 \\

\addlinespace[2pt]

TRPO-Lag & $(0.03,0.00)$ & 2811.96 & 2511.97 & 36.21 & 51.21 \\
TRPO-Lag & $(0.03,0.03)$ & 2724.20 & 2424.20 & 42.56 & 57.56 \\
TRPO-Lag & $(0.03,0.06)$ & 2446.67 & 2146.68 & 45.14 & 60.14 \\

\addlinespace[2pt]

TRPO-Lag & $(0.06,0.00)$ & 2546.27 & 2246.28 & 42.33 & 57.33 \\
TRPO-Lag & $(0.06,0.03)$ & 2456.10 & 2156.11 & 42.98 & 57.98 \\
TRPO-Lag & $(0.06,0.06)$ & 2292.36 & 1992.37 & 39.32 & 54.32 \\

\bottomrule
\end{tabular*}
\end{table}

\begin{table}[!htbp]
\centering
\caption{Reliability certificates for TRPO-Lagrangian on SafetyAnt under bounded $L_\infty$ perturbations.}
\label{tab:ant-linf-trpolag-rel}

\small
\renewcommand{\arraystretch}{1.12}
\setlength{\tabcolsep}{5pt}

\begin{tabular*}{\textwidth}{@{\extracolsep{\fill}} l c r r r r c}
\toprule
\textbf{Policy}
& \textbf{$(k_a,k_s)$}
& \textbf{$\hat p_{\mathrm{task}}$}
& \textbf{$\hat p_{\mathrm{safety}}$}
& \textbf{$\hat p_F$}
& \textbf{B-CP UB}
& \textbf{Decision} \\
\midrule

TRPO-Lag & $(0.00,0.00)$ & 0.0076 & 0.0000 & 0.0076 & \textbf{0.0098} & \textbf{PASS} \\
TRPO-Lag & $(0.00,0.03)$ & 0.0092 & 0.0001 & 0.0093 & \textbf{0.0117} & \textbf{PASS} \\
TRPO-Lag & $(0.00,0.06)$ & 0.0265 & 0.0048 & 0.0313 & \textbf{0.0355} & \textbf{PASS} \\

\addlinespace[2pt]

TRPO-Lag & $(0.03,0.00)$ & 0.0097 & 0.0001 & 0.0098 & \textbf{0.0123} & \textbf{PASS} \\
TRPO-Lag & $(0.03,0.03)$ & 0.0163 & 0.0009 & 0.0172 & \textbf{0.0204} & \textbf{PASS} \\
TRPO-Lag & $(0.03,0.06)$ & 0.0339 & 0.0025 & 0.0364 & \textbf{0.0409} & \textbf{PASS} \\

\addlinespace[2pt]

TRPO-Lag & $(0.06,0.00)$ & 0.0338 & 0.0007 & 0.0345 & \textbf{0.0389} & \textbf{PASS} \\
TRPO-Lag & $(0.06,0.03)$ & 0.0455 & 0.0009 & 0.0464 & 0.0514 & FAIL \\
TRPO-Lag & $(0.06,0.06)$ & 0.0621 & 0.0014 & 0.0635 & 0.0693 & FAIL \\

\bottomrule
\end{tabular*}
\end{table}

\FloatBarrier






\section{Complete Results for Safety Point Goal}
\label{App:point}


\FloatBarrier
\subsection{Gaussian Noise}

The Gaussian perturbation experiments evaluate state and action noise over
$k_a,k_s\in\{0.00,0.05,0.10\}$. Each policy--configuration pair is
evaluated using $N=10{,}000$ episodes. Certification uses a Bonferroni
family size of $M=9$ per policy, fixed safety budget $d=100$, and
deployment failure threshold $p^\star=0.05$. For SafetyPointGoal,
task failure corresponds to failure to reach the goal.











\begin{table}[!htbp]
\centering
\caption{Reliability certificates for PPO on SafetyPointGoal under Gaussian perturbations.}
\label{tab:point-gaussian-ppo-rel}

\small
\renewcommand{\arraystretch}{1.12}
\setlength{\tabcolsep}{5pt}

\begin{tabular*}{\textwidth}{@{\extracolsep{\fill}} l c r r r r c}
\toprule
\textbf{Policy}
& \textbf{$(k_a,k_s)$}
& \textbf{$\hat p_{\mathrm{task}}$}
& \textbf{$\hat p_{\mathrm{safety}}$}
& \textbf{$\hat p_F$}
& \textbf{B-CP UB}
& \textbf{Decision} \\
\midrule

PPO & $(0.00,0.00)$ & 0.0000 & 0.1060 & 0.1060 & 0.1132 & FAIL \\
PPO & $(0.00,0.05)$ & 0.0476 & 0.2561 & 0.2898 & 0.3003 & FAIL \\
PPO & $(0.00,0.10)$ & 0.3093 & 0.2324 & 0.4583 & 0.4698 & FAIL \\

\addlinespace[2pt]

PPO & $(0.05,0.00)$ & 0.0000 & 0.1090 & 0.1090 & 0.1163 & FAIL \\
PPO & $(0.05,0.05)$ & 0.0450 & 0.2612 & 0.2905 & 0.3010 & FAIL \\
PPO & $(0.05,0.10)$ & 0.3154 & 0.2413 & 0.4697 & 0.4812 & FAIL \\

\addlinespace[2pt]

PPO & $(0.10,0.00)$ & 0.0000 & 0.1037 & 0.1037 & 0.1109 & FAIL \\
PPO & $(0.10,0.05)$ & 0.0546 & 0.2600 & 0.2979 & 0.3085 & FAIL \\
PPO & $(0.10,0.10)$ & 0.3378 & 0.2355 & 0.4790 & 0.4905 & FAIL \\

\bottomrule
\end{tabular*}
\end{table}

\FloatBarrier











\begin{table}[!htbp]
\centering
\caption{Reliability certificates for PPO-Lagrangian on SafetyPointGoal under Gaussian perturbations.}
\label{tab:point-gaussian-ppolag-rel}

\small
\renewcommand{\arraystretch}{1.12}
\setlength{\tabcolsep}{5pt}

\begin{tabular*}{\textwidth}{@{\extracolsep{\fill}} l c r r r r c}
\toprule
\textbf{Policy}
& \textbf{$(k_a,k_s)$}
& \textbf{$\hat p_{\mathrm{task}}$}
& \textbf{$\hat p_{\mathrm{safety}}$}
& \textbf{$\hat p_F$}
& \textbf{B-CP UB}
& \textbf{Decision} \\
\midrule

PPO-Lag & $(0.00,0.00)$ & 0.0580 & 0.0202 & 0.0775 & 0.0838 & FAIL \\
PPO-Lag & $(0.00,0.05)$ & 0.4898 & 0.1745 & 0.5747 & 0.5860 & FAIL \\
PPO-Lag & $(0.00,0.10)$ & 0.7497 & 0.1662 & 0.7858 & 0.7951 & FAIL \\

\addlinespace[2pt]

PPO-Lag & $(0.05,0.00)$ & 0.0574 & 0.0248 & 0.0809 & 0.0873 & FAIL \\
PPO-Lag & $(0.05,0.05)$ & 0.5030 & 0.1655 & 0.5774 & 0.5887 & FAIL \\
PPO-Lag & $(0.05,0.10)$ & 0.7542 & 0.1751 & 0.7927 & 0.8019 & FAIL \\

\addlinespace[2pt]

PPO-Lag & $(0.10,0.00)$ & 0.0587 & 0.0234 & 0.0810 & 0.0875 & FAIL \\
PPO-Lag & $(0.10,0.05)$ & 0.5058 & 0.1652 & 0.5814 & 0.5927 & FAIL \\
PPO-Lag & $(0.10,0.10)$ & 0.7484 & 0.1720 & 0.7865 & 0.7958 & FAIL \\

\bottomrule
\end{tabular*}
\end{table}

\FloatBarrier











\begin{table}[!htbp]
\centering
\caption{Reliability certificates for TRPO on SafetyPointGoal under Gaussian perturbations.}
\label{tab:point-gaussian-trpo-rel}

\small
\renewcommand{\arraystretch}{1.12}
\setlength{\tabcolsep}{5pt}

\begin{tabular*}{\textwidth}{@{\extracolsep{\fill}} l c r r r r c}
\toprule
\textbf{Policy}
& \textbf{$(k_a,k_s)$}
& \textbf{$\hat p_{\mathrm{task}}$}
& \textbf{$\hat p_{\mathrm{safety}}$}
& \textbf{$\hat p_F$}
& \textbf{B-CP UB}
& \textbf{Decision} \\
\midrule

TRPO & $(0.00,0.00)$ & 0.0000 & 0.1228 & 0.1228 & 0.1305 & FAIL \\
TRPO & $(0.00,0.05)$ & 0.0207 & 0.2485 & 0.2628 & 0.2730 & FAIL \\
TRPO & $(0.00,0.10)$ & 0.2058 & 0.2258 & 0.3761 & 0.3873 & FAIL \\

\addlinespace[2pt]

TRPO & $(0.05,0.00)$ & 0.0000 & 0.1213 & 0.1213 & 0.1290 & FAIL \\
TRPO & $(0.05,0.05)$ & 0.0238 & 0.2515 & 0.2686 & 0.2789 & FAIL \\
TRPO & $(0.05,0.10)$ & 0.2043 & 0.2282 & 0.3768 & 0.3880 & FAIL \\

\addlinespace[2pt]

TRPO & $(0.10,0.00)$ & 0.0000 & 0.1181 & 0.1181 & 0.1257 & FAIL \\
TRPO & $(0.10,0.05)$ & 0.0235 & 0.2517 & 0.2696 & 0.2799 & FAIL \\
TRPO & $(0.10,0.10)$ & 0.2164 & 0.2194 & 0.3792 & 0.3904 & FAIL \\

\bottomrule
\end{tabular*}
\end{table}

\FloatBarrier


\begin{table}[!htbp]
\centering
\caption{Performance and cost certificates for TRPO-Lagrangian on SafetyPointGoal under Gaussian perturbations.}
\label{tab:point-gaussian-trpolag-perf}

\small
\renewcommand{\arraystretch}{1.12}
\setlength{\tabcolsep}{5pt}

\begin{tabular*}{\textwidth}{@{\extracolsep{\fill}} l c r r r r}
\toprule
\textbf{Policy}
& \textbf{$(k_a,k_s)$}
& \textbf{$\bar{R}$}
& \textbf{$R_{\mathrm{H\text{-}LCB}}$}
& \textbf{$\bar{C}$}
& \textbf{$C_{\mathrm{H\text{-}UCB}}$} \\
\midrule

TRPO-Lag & $(0.00,0.00)$ & 21.72 & -278.27 & 29.28 & 44.28 \\
TRPO-Lag & $(0.00,0.05)$ & 2.25 & -297.74 & 31.82 & 46.82 \\
TRPO-Lag & $(0.00,0.10)$ & 0.97 & -299.02 & 39.67 & 54.67 \\

\addlinespace[2pt]

TRPO-Lag & $(0.05,0.00)$ & 21.72 & -278.27 & 29.95 & 44.95 \\
TRPO-Lag & $(0.05,0.05)$ & 2.24 & -297.75 & 32.65 & 47.65 \\
TRPO-Lag & $(0.05,0.10)$ & 0.96 & -299.03 & 40.84 & 55.84 \\

\addlinespace[2pt]

TRPO-Lag & $(0.10,0.00)$ & 21.60 & -278.39 & 28.97 & 43.97 \\
TRPO-Lag & $(0.10,0.05)$ & 2.22 & -297.78 & 32.33 & 47.33 \\
TRPO-Lag & $(0.10,0.10)$ & 0.95 & -299.04 & 40.77 & 55.77 \\

\bottomrule
\end{tabular*}
\end{table}

\begin{table}[!htbp]
\centering
\caption{Reliability certificates for TRPO-Lagrangian on SafetyPointGoal under Gaussian perturbations.}
\label{tab:point-gaussian-trpolag-rel}

\small
\renewcommand{\arraystretch}{1.12}
\setlength{\tabcolsep}{5pt}

\begin{tabular*}{\textwidth}{@{\extracolsep{\fill}} l c r r r r c}
\toprule
\textbf{Policy}
& \textbf{$(k_a,k_s)$}
& \textbf{$\hat p_{\mathrm{task}}$}
& \textbf{$\hat p_{\mathrm{safety}}$}
& \textbf{$\hat p_F$}
& \textbf{B-CP UB}
& \textbf{Decision} \\
\midrule

TRPO-Lag & $(0.00,0.00)$ & 0.0080 & 0.0209 & 0.0287 &
\textbf{0.0328} & \textbf{PASS} \\
TRPO-Lag & $(0.00,0.05)$ & 0.3095 & 0.1120 & 0.3889 & 0.4001 & FAIL \\
TRPO-Lag & $(0.00,0.10)$ & 0.6201 & 0.1516 & 0.6739 & 0.6846 & FAIL \\

\addlinespace[2pt]

TRPO-Lag & $(0.05,0.00)$ & 0.0090 & 0.0222 & 0.0310 &
\textbf{0.0352} & \textbf{PASS} \\
TRPO-Lag & $(0.05,0.05)$ & 0.3156 & 0.1197 & 0.3965 & 0.4078 & FAIL \\
TRPO-Lag & $(0.05,0.10)$ & 0.6196 & 0.1558 & 0.6728 & 0.6835 & FAIL \\

\addlinespace[2pt]

TRPO-Lag & $(0.10,0.00)$ & 0.0087 & 0.0214 & 0.0296 &
\textbf{0.0337} & \textbf{PASS} \\
TRPO-Lag & $(0.10,0.05)$ & 0.3173 & 0.1153 & 0.3955 & 0.4068 & FAIL \\
TRPO-Lag & $(0.10,0.10)$ & 0.6305 & 0.1509 & 0.6815 & 0.6921 & FAIL \\

\bottomrule
\end{tabular*}
\end{table}

\FloatBarrier








\subsection{Uniform Noise}

The Uniform perturbation experiments evaluate state and action noise over
$k_a,k_s\in\{0.00,0.05,0.10\}$. Each policy--configuration pair is
evaluated using $N=10{,}000$ episodes. Certification uses a Bonferroni
family size of $M=9$ per policy, fixed safety budget $d=100$, and
deployment failure threshold $p^\star=0.05$. For SafetyPointGoal,
task failure corresponds to failure to reach the goal.











\begin{table}[!htbp]
\centering
\caption{Reliability certificates for PPO on SafetyPointGoal under Uniform perturbations.}
\label{tab:point-uniform-ppo-rel}

\small
\renewcommand{\arraystretch}{1.12}
\setlength{\tabcolsep}{5pt}

\begin{tabular*}{\textwidth}{@{\extracolsep{\fill}} l c r r r r c}
\toprule
\textbf{Policy}
& \textbf{$(k_a,k_s)$}
& \textbf{$\hat p_{\mathrm{task}}$}
& \textbf{$\hat p_{\mathrm{safety}}$}
& \textbf{$\hat p_F$}
& \textbf{B-CP UB}
& \textbf{Decision} \\
\midrule

PPO & $(0.00,0.00)$ & 0.0000 & 0.1060 & 0.1060 & 0.1132 & FAIL \\
PPO & $(0.00,0.05)$ & 0.0126 & 0.2468 & 0.2547 & 0.2648 & FAIL \\
PPO & $(0.00,0.10)$ & 0.0846 & 0.2598 & 0.3177 & 0.3285 & FAIL \\

\addlinespace[2pt]

PPO & $(0.05,0.00)$ & 0.0001 & 0.1086 & 0.1086 & 0.1159 & FAIL \\
PPO & $(0.05,0.05)$ & 0.0140 & 0.2401 & 0.2496 & 0.2596 & FAIL \\
PPO & $(0.05,0.10)$ & 0.0886 & 0.2560 & 0.3171 & 0.3279 & FAIL \\

\addlinespace[2pt]

PPO & $(0.10,0.00)$ & 0.0000 & 0.1081 & 0.1081 & 0.1154 & FAIL \\
PPO & $(0.10,0.05)$ & 0.0140 & 0.2406 & 0.2502 & 0.2602 & FAIL \\
PPO & $(0.10,0.10)$ & 0.0955 & 0.2567 & 0.3221 & 0.3329 & FAIL \\

\bottomrule
\end{tabular*}
\end{table}

\FloatBarrier











\begin{table}[!htbp]
\centering
\caption{Reliability certificates for PPO-Lagrangian on SafetyPointGoal under Uniform perturbations.}
\label{tab:point-uniform-ppolag-rel}

\small
\renewcommand{\arraystretch}{1.12}
\setlength{\tabcolsep}{5pt}

\begin{tabular*}{\textwidth}{@{\extracolsep{\fill}} l c r r r r c}
\toprule
\textbf{Policy}
& \textbf{$(k_a,k_s)$}
& \textbf{$\hat p_{\mathrm{task}}$}
& \textbf{$\hat p_{\mathrm{safety}}$}
& \textbf{$\hat p_F$}
& \textbf{B-CP UB}
& \textbf{Decision} \\
\midrule

PPO-Lag & $(0.00,0.00)$ & 0.0580 & 0.0202 & 0.0775 & 0.0838 & FAIL \\
PPO-Lag & $(0.00,0.05)$ & 0.2994 & 0.1553 & 0.4052 & 0.4165 & FAIL \\
PPO-Lag & $(0.00,0.10)$ & 0.5791 & 0.1680 & 0.6401 & 0.6511 & FAIL \\

\addlinespace[2pt]

PPO-Lag & $(0.05,0.00)$ & 0.0611 & 0.0226 & 0.0821 & 0.0886 & FAIL \\
PPO-Lag & $(0.05,0.05)$ & 0.3019 & 0.1648 & 0.4129 & 0.4242 & FAIL \\
PPO-Lag & $(0.05,0.10)$ & 0.5763 & 0.1770 & 0.6403 & 0.6513 & FAIL \\

\addlinespace[2pt]

PPO-Lag & $(0.10,0.00)$ & 0.0598 & 0.0223 & 0.0800 & 0.0864 & FAIL \\
PPO-Lag & $(0.10,0.05)$ & 0.3071 & 0.1604 & 0.4145 & 0.4258 & FAIL \\
PPO-Lag & $(0.10,0.10)$ & 0.5818 & 0.1754 & 0.6472 & 0.6581 & FAIL \\

\bottomrule
\end{tabular*}
\end{table}

\FloatBarrier











\begin{table}[!htbp]
\centering
\caption{Reliability certificates for TRPO on SafetyPointGoal under Uniform perturbations.}
\label{tab:point-uniform-trpo-rel}

\small
\renewcommand{\arraystretch}{1.12}
\setlength{\tabcolsep}{5pt}

\begin{tabular*}{\textwidth}{@{\extracolsep{\fill}} l c r r r r c}
\toprule
\textbf{Policy}
& \textbf{$(k_a,k_s)$}
& \textbf{$\hat p_{\mathrm{task}}$}
& \textbf{$\hat p_{\mathrm{safety}}$}
& \textbf{$\hat p_F$}
& \textbf{B-CP UB}
& \textbf{Decision} \\
\midrule

TRPO & $(0.00,0.00)$ & 0.0000 & 0.1228 & 0.1228 & 0.1305 & FAIL \\
TRPO & $(0.00,0.05)$ & 0.0055 & 0.2372 & 0.2408 & 0.2507 & FAIL \\
TRPO & $(0.00,0.10)$ & 0.0442 & 0.2435 & 0.2769 & 0.2873 & FAIL \\

\addlinespace[2pt]

TRPO & $(0.05,0.00)$ & 0.0000 & 0.1202 & 0.1202 & 0.1278 & FAIL \\
TRPO & $(0.05,0.05)$ & 0.0052 & 0.2297 & 0.2335 & 0.2433 & FAIL \\
TRPO & $(0.05,0.10)$ & 0.0429 & 0.2451 & 0.2744 & 0.2847 & FAIL \\

\addlinespace[2pt]

TRPO & $(0.10,0.00)$ & 0.0000 & 0.1212 & 0.1212 & 0.1289 & FAIL \\
TRPO & $(0.10,0.05)$ & 0.0050 & 0.2332 & 0.2367 & 0.2466 & FAIL \\
TRPO & $(0.10,0.10)$ & 0.0442 & 0.2488 & 0.2799 & 0.2903 & FAIL \\

\bottomrule
\end{tabular*}
\end{table}

\FloatBarrier


\begin{table}[!htbp]
\centering
\caption{Performance and cost certificates for TRPO-Lagrangian on SafetyPointGoal under Uniform perturbations.}
\label{tab:point-uniform-trpolag-perf}

\small
\renewcommand{\arraystretch}{1.12}
\setlength{\tabcolsep}{5pt}

\begin{tabular*}{\textwidth}{@{\extracolsep{\fill}} l c r r r r}
\toprule
\textbf{Policy}
& \textbf{$(k_a,k_s)$}
& \textbf{$\bar{R}$}
& \textbf{$R_{\mathrm{H\text{-}LCB}}$}
& \textbf{$\bar{C}$}
& \textbf{$C_{\mathrm{H\text{-}UCB}}$} \\
\midrule

TRPO-Lag & $(0.00,0.00)$ & 21.72 & -278.27 & 29.28 & 44.28 \\
TRPO-Lag & $(0.00,0.05)$ & 3.57 & -296.42 & 26.13 & 41.13 \\
TRPO-Lag & $(0.00,0.10)$ & 1.75 & -298.24 & 35.97 & 50.97 \\

\addlinespace[2pt]

TRPO-Lag & $(0.05,0.00)$ & 21.75 & -278.25 & 30.10 & 45.10 \\
TRPO-Lag & $(0.05,0.05)$ & 3.58 & -296.41 & 25.56 & 40.56 \\
TRPO-Lag & $(0.05,0.10)$ & 1.74 & -298.26 & 34.79 & 49.79 \\

\addlinespace[2pt]

TRPO-Lag & $(0.10,0.00)$ & 21.67 & -278.32 & 29.31 & 44.31 \\
TRPO-Lag & $(0.10,0.05)$ & 3.57 & -296.42 & 25.18 & 40.18 \\
TRPO-Lag & $(0.10,0.10)$ & 1.73 & -298.26 & 35.30 & 50.30 \\

\bottomrule
\end{tabular*}
\end{table}

\begin{table}[!htbp]
\centering
\caption{Reliability certificates for TRPO-Lagrangian on SafetyPointGoal under Uniform perturbations.}
\label{tab:point-uniform-trpolag-rel}

\small
\renewcommand{\arraystretch}{1.12}
\setlength{\tabcolsep}{5pt}

\begin{tabular*}{\textwidth}{@{\extracolsep{\fill}} l c r r r r c}
\toprule
\textbf{Policy}
& \textbf{$(k_a,k_s)$}
& \textbf{$\hat p_{\mathrm{task}}$}
& \textbf{$\hat p_{\mathrm{safety}}$}
& \textbf{$\hat p_F$}
& \textbf{B-CP UB}
& \textbf{Decision} \\
\midrule

TRPO-Lag & $(0.00,0.00)$ & 0.0080 & 0.0209 & 0.0287 &
\textbf{0.0328} & \textbf{PASS} \\
TRPO-Lag & $(0.00,0.05)$ & 0.1818 & 0.0824 & 0.2523 & 0.2624 & FAIL \\
TRPO-Lag & $(0.00,0.10)$ & 0.4071 & 0.1307 & 0.4830 & 0.4945 & FAIL \\

\addlinespace[2pt]

TRPO-Lag & $(0.05,0.00)$ & 0.0084 & 0.0219 & 0.0300 &
\textbf{0.0341} & \textbf{PASS} \\
TRPO-Lag & $(0.05,0.05)$ & 0.1791 & 0.0811 & 0.2486 & 0.2586 & FAIL \\
TRPO-Lag & $(0.05,0.10)$ & 0.4120 & 0.1287 & 0.4871 & 0.4986 & FAIL \\

\addlinespace[2pt]

TRPO-Lag & $(0.10,0.00)$ & 0.0087 & 0.0210 & 0.0293 &
\textbf{0.0334} & \textbf{PASS} \\
TRPO-Lag & $(0.10,0.05)$ & 0.1824 & 0.0798 & 0.2488 & 0.2588 & FAIL \\
TRPO-Lag & $(0.10,0.10)$ & 0.4087 & 0.1293 & 0.4830 & 0.4945 & FAIL \\

\bottomrule
\end{tabular*}
\end{table}

\FloatBarrier

\subsection{Bounded Adversarial $L_\infty$ Noise}

The bounded $L_\infty$ perturbation experiments evaluate state and action
perturbations over $k_a,k_s\in\{0.00,0.03,0.06\}$. Each
policy--configuration pair is evaluated using $N=10{,}000$ episodes.
Certification uses a Bonferroni family size of $M=9$ per policy, fixed
safety budget $d=100$, and deployment failure threshold $p^\star=0.05$.
For SafetyPointGoal, task failure corresponds to failure to reach the goal.











\begin{table}[!htbp]
\centering
\caption{Reliability certificates for PPO on SafetyPointGoal under bounded $L_\infty$ perturbations.}
\label{tab:point-linf-ppo-rel}

\small
\renewcommand{\arraystretch}{1.12}
\setlength{\tabcolsep}{5pt}

\begin{tabular*}{\textwidth}{@{\extracolsep{\fill}} l c r r r r c}
\toprule
\textbf{Policy}
& \textbf{$(k_a,k_s)$}
& \textbf{$\hat p_{\mathrm{task}}$}
& \textbf{$\hat p_{\mathrm{safety}}$}
& \textbf{$\hat p_F$}
& \textbf{B-CP UB}
& \textbf{Decision} \\
\midrule

PPO & $(0.00,0.00)$ & 0.0000 & 0.1060 & 0.1060 & 0.1132 & FAIL \\
PPO & $(0.00,0.03)$ & 0.0000 & 0.1108 & 0.1108 & 0.1182 & FAIL \\
PPO & $(0.00,0.06)$ & 0.0000 & 0.1058 & 0.1058 & 0.1130 & FAIL \\

\addlinespace[2pt]

PPO & $(0.03,0.00)$ & 0.0000 & 0.1056 & 0.1056 & 0.1128 & FAIL \\
PPO & $(0.03,0.03)$ & 0.0000 & 0.1085 & 0.1085 & 0.1158 & FAIL \\
PPO & $(0.03,0.06)$ & 0.0000 & 0.1058 & 0.1058 & 0.1130 & FAIL \\

\addlinespace[2pt]

PPO & $(0.06,0.00)$ & 0.0000 & 0.1086 & 0.1086 & 0.1159 & FAIL \\
PPO & $(0.06,0.03)$ & 0.0000 & 0.1089 & 0.1089 & 0.1162 & FAIL \\
PPO & $(0.06,0.06)$ & 0.0001 & 0.1082 & 0.1082 & 0.1155 & FAIL \\

\bottomrule
\end{tabular*}
\end{table}

\FloatBarrier











\begin{table}[!htbp]
\centering
\caption{Reliability certificates for PPO-Lagrangian on SafetyPointGoal under bounded $L_\infty$ perturbations.}
\label{tab:point-linf-ppolag-rel}

\small
\renewcommand{\arraystretch}{1.12}
\setlength{\tabcolsep}{5pt}

\begin{tabular*}{\textwidth}{@{\extracolsep{\fill}} l c r r r r c}
\toprule
\textbf{Policy}
& \textbf{$(k_a,k_s)$}
& \textbf{$\hat p_{\mathrm{task}}$}
& \textbf{$\hat p_{\mathrm{safety}}$}
& \textbf{$\hat p_F$}
& \textbf{B-CP UB}
& \textbf{Decision} \\
\midrule

PPO-Lag & $(0.00,0.00)$ & 0.0580 & 0.0202 & 0.0775 & 0.0838 & FAIL \\
PPO-Lag & $(0.00,0.03)$ & 0.0580 & 0.0244 & 0.0815 & 0.0880 & FAIL \\
PPO-Lag & $(0.00,0.06)$ & 0.0569 & 0.0216 & 0.0770 & 0.0833 & FAIL \\

\addlinespace[2pt]

PPO-Lag & $(0.03,0.00)$ & 0.0600 & 0.0201 & 0.0792 & 0.0856 & FAIL \\
PPO-Lag & $(0.03,0.03)$ & 0.0579 & 0.0223 & 0.0788 & 0.0852 & FAIL \\
PPO-Lag & $(0.03,0.06)$ & 0.0575 & 0.0199 & 0.0765 & 0.0828 & FAIL \\

\addlinespace[2pt]

PPO-Lag & $(0.06,0.00)$ & 0.0585 & 0.0210 & 0.0781 & 0.0844 & FAIL \\
PPO-Lag & $(0.06,0.03)$ & 0.0586 & 0.0203 & 0.0776 & 0.0839 & FAIL \\
PPO-Lag & $(0.06,0.06)$ & 0.0572 & 0.0209 & 0.0773 & 0.0836 & FAIL \\

\bottomrule
\end{tabular*}
\end{table}

\FloatBarrier











\begin{table}[!htbp]
\centering
\caption{Reliability certificates for TRPO on SafetyPointGoal under bounded $L_\infty$ perturbations.}
\label{tab:point-linf-trpo-rel}

\small
\renewcommand{\arraystretch}{1.12}
\setlength{\tabcolsep}{5pt}

\begin{tabular*}{\textwidth}{@{\extracolsep{\fill}} l c r r r r c}
\toprule
\textbf{Policy}
& \textbf{$(k_a,k_s)$}
& \textbf{$\hat p_{\mathrm{task}}$}
& \textbf{$\hat p_{\mathrm{safety}}$}
& \textbf{$\hat p_F$}
& \textbf{B-CP UB}
& \textbf{Decision} \\
\midrule

TRPO & $(0.00,0.00)$ & 0.0000 & 0.1228 & 0.1228 & 0.1305 & FAIL \\
TRPO & $(0.00,0.03)$ & 0.0000 & 0.1190 & 0.1190 & 0.1266 & FAIL \\
TRPO & $(0.00,0.06)$ & 0.0000 & 0.1201 & 0.1201 & 0.1277 & FAIL \\

\addlinespace[2pt]

TRPO & $(0.03,0.00)$ & 0.0001 & 0.1219 & 0.1219 & 0.1296 & FAIL \\
TRPO & $(0.03,0.03)$ & 0.0000 & 0.1196 & 0.1196 & 0.1272 & FAIL \\
TRPO & $(0.03,0.06)$ & 0.0000 & 0.1164 & 0.1164 & 0.1239 & FAIL \\

\addlinespace[2pt]

TRPO & $(0.06,0.00)$ & 0.0000 & 0.1153 & 0.1153 & 0.1228 & FAIL \\
TRPO & $(0.06,0.03)$ & 0.0000 & 0.1184 & 0.1184 & 0.1260 & FAIL \\
TRPO & $(0.06,0.06)$ & 0.0000 & 0.1161 & 0.1161 & 0.1236 & FAIL \\

\bottomrule
\end{tabular*}
\end{table}

\FloatBarrier


\begin{table}[!htbp]
\centering
\caption{Performance and cost certificates for TRPO-Lagrangian on SafetyPointGoal under bounded $L_\infty$ perturbations.}
\label{tab:point-linf-trpolag-perf}

\small
\renewcommand{\arraystretch}{1.12}
\setlength{\tabcolsep}{5pt}

\begin{tabular*}{\textwidth}{@{\extracolsep{\fill}} l c r r r r}
\toprule
\textbf{Policy}
& \textbf{$(k_a,k_s)$}
& \textbf{$\bar{R}$}
& \textbf{$R_{\mathrm{H\text{-}LCB}}$}
& \textbf{$\bar{C}$}
& \textbf{$C_{\mathrm{H\text{-}UCB}}$} \\
\midrule

TRPO-Lag & $(0.00,0.00)$ & 21.72 & -278.27 & 29.28 & 44.28 \\
TRPO-Lag & $(0.00,0.03)$ & 21.77 & -278.23 & 29.42 & 44.42 \\
TRPO-Lag & $(0.00,0.06)$ & 21.85 & -278.14 & 29.74 & 44.74 \\

\addlinespace[2pt]

TRPO-Lag & $(0.03,0.00)$ & 21.86 & -278.13 & 29.35 & 44.35 \\
TRPO-Lag & $(0.03,0.03)$ & 21.76 & -278.23 & 30.06 & 45.06 \\
TRPO-Lag & $(0.03,0.06)$ & 21.77 & -278.22 & 29.23 & 44.23 \\

\addlinespace[2pt]

TRPO-Lag & $(0.06,0.00)$ & 21.73 & -278.27 & 29.55 & 44.54 \\
TRPO-Lag & $(0.06,0.03)$ & 21.73 & -278.26 & 29.58 & 44.58 \\
TRPO-Lag & $(0.06,0.06)$ & 21.77 & -278.22 & 28.96 & 43.96 \\

\bottomrule
\end{tabular*}
\end{table}

\begin{table}[!htbp]
\centering
\caption{Reliability certificates for TRPO-Lagrangian on SafetyPointGoal under bounded $L_\infty$ perturbations.}
\label{tab:point-linf-trpolag-rel}

\small
\renewcommand{\arraystretch}{1.12}
\setlength{\tabcolsep}{5pt}

\begin{tabular*}{\textwidth}{@{\extracolsep{\fill}} l c r r r r c}
\toprule
\textbf{Policy}
& \textbf{$(k_a,k_s)$}
& \textbf{$\hat p_{\mathrm{task}}$}
& \textbf{$\hat p_{\mathrm{safety}}$}
& \textbf{$\hat p_F$}
& \textbf{B-CP UB}
& \textbf{Decision} \\
\midrule

TRPO-Lag & $(0.00,0.00)$ & 0.0080 & 0.0209 & 0.0287 &
\textbf{0.0328} & \textbf{PASS} \\
TRPO-Lag & $(0.00,0.03)$ & 0.0086 & 0.0187 & 0.0271 &
\textbf{0.0310} & \textbf{PASS} \\
TRPO-Lag & $(0.00,0.06)$ & 0.0078 & 0.0211 & 0.0286 &
\textbf{0.0326} & \textbf{PASS} \\

\addlinespace[2pt]

TRPO-Lag & $(0.03,0.00)$ & 0.0084 & 0.0217 & 0.0297 &
\textbf{0.0338} & \textbf{PASS} \\
TRPO-Lag & $(0.03,0.03)$ & 0.0085 & 0.0216 & 0.0297 &
\textbf{0.0338} & \textbf{PASS} \\
TRPO-Lag & $(0.03,0.06)$ & 0.0083 & 0.0229 & 0.0310 &
\textbf{0.0352} & \textbf{PASS} \\

\addlinespace[2pt]

TRPO-Lag & $(0.06,0.00)$ & 0.0085 & 0.0192 & 0.0274 &
\textbf{0.0314} & \textbf{PASS} \\
TRPO-Lag & $(0.06,0.03)$ & 0.0089 & 0.0223 & 0.0308 &
\textbf{0.0350} & \textbf{PASS} \\
TRPO-Lag & $(0.06,0.06)$ & 0.0084 & 0.0204 & 0.0285 &
\textbf{0.0325} & \textbf{PASS} \\

\bottomrule
\end{tabular*}
\end{table}

\FloatBarrier









\FloatBarrier

\section{Sensitivity on Safety Thresholds}
\label{App:Sens}
\begin{table}[!h]
\caption{TRPO-Lagrangian reliability certificates under varying safety-budget
thresholds $d$. Each entry reports the mean Bonferroni-corrected
Clopper--Pearson upper bound $B_{\mathrm{CP}}^{\mathrm{UB}}$ across the nine
$(k_a,k_s)$ perturbation configurations.}
\label{tab:trpolag_d_bcpub}
\centering
\small
\setlength{\tabcolsep}{4pt}

\resizebox{\textwidth}{!}{%
\begin{tabular}{llcccccccc}
\toprule
\textbf{Environment}
& \textbf{Perturbation}
& $d=25$
& $d=50$
& $d=100$
& $d=250$
& $d=500$
& $d=750$
& $d=1000$
& $d=1500$ \\
\midrule

\multirow{3}{*}{SafetyAnt}
& Gaussian
& 0.8361 & 0.7085 & 0.6766 & 0.6765
& 0.6765 & 0.6765 & 0.6765 & 0.6765 \\

& Uniform
& 0.8547 & 0.7100 & 0.6404 & 0.6402
& 0.6402 & 0.6402 & 0.6402 & 0.6402 \\

& $L_\infty$
& 0.7952 & 0.3762 & 0.0323 & 0.0309
& 0.0309 & 0.0309 & 0.0309 & 0.0309 \\

\midrule

\multirow{3}{*}{SafetyCar}
& Gaussian
& 0.7337 & 0.6359 & 0.5503 & 0.5173
& 0.5150 & 0.5149 & 0.5149 & 0.5149 \\

& Uniform
& 0.6740 & 0.5647 & 0.4703 & 0.4295
& 0.4252 & 0.4250 & 0.4250 & 0.4250 \\

& $L_\infty$
& 0.4639 & 0.2196 & 0.0515 & 0.0229
& 0.0200 & 0.0194 & 0.0194 & 0.0194 \\

\midrule

\multirow{3}{*}{SafetyPointGoal}
& Gaussian
& 0.5791 & 0.4627 & 0.3752 & 0.3334
& 0.3242 & 0.3235 & 0.3234 & 0.3234 \\

& Uniform
& 0.4997 & 0.3630 & 0.2631 & 0.2169
& 0.2080 & 0.2072 & 0.2071 & 0.2071 \\

& $L_\infty$
& 0.4750 & 0.1991 & 0.0331 & 0.0147
& 0.0119 & 0.0109 & 0.0107 & 0.0107 \\

\bottomrule
\end{tabular}%
}
\end{table}
\begin{table}[!h]
\caption{TRPO-Lagrangian reliability certification under varying reliability
thresholds $p^\star$. The Bonferroni-corrected Clopper--Pearson upper bound
$B_{\mathrm{CP}}^{\mathrm{UB}}$ is reported for each environment and
perturbation type. A certificate passes when
$B_{\mathrm{CP}}^{\mathrm{UB}} \leq p^\star$.}
\label{tab:trpolag_pstar_sweep}
\centering
\small
\setlength{\tabcolsep}{5pt}

\resizebox{\textwidth}{!}{%
\begin{tabular}{llc|cccc}
\toprule
\textbf{Environment}
& \textbf{Perturbation}
& $B_{\mathrm{CP}}^{\mathrm{UB}}$
& $p^\star=0.010$
& $p^\star=0.025$
& $p^\star=0.050$
& $p^\star=0.100$ \\
\midrule

\multirow{3}{*}{SafetyAnt}
& Gaussian
& 0.677
& FAIL & FAIL & FAIL & FAIL \\

& Uniform
& 0.640
& FAIL & FAIL & FAIL & FAIL \\

& $L_\infty$
& 0.032
& FAIL & FAIL & PASS & PASS \\

\midrule

\multirow{3}{*}{SafetyCar}
& Gaussian
& 0.550
& FAIL & FAIL & FAIL & FAIL \\

& Uniform
& 0.470
& FAIL & FAIL & FAIL & FAIL \\

& $L_\infty$
& 0.052
& FAIL & FAIL & FAIL & PASS \\

\midrule

\multirow{3}{*}{SafetyPointGoal}
& Gaussian
& 0.375
& FAIL & FAIL & FAIL & FAIL \\

& Uniform
& 0.263
& FAIL & FAIL & FAIL & FAIL \\

& $L_\infty$
& 0.033
& FAIL & FAIL & PASS & PASS \\

\bottomrule
\end{tabular}%
}
\end{table}
\endgroup
\section{Sample Complexity for Deployment-Grade Certification}
\label{app:sample_complexity}

A key practical consideration in certification is determining the number of evaluation episodes $N$ required to certify a target reliability threshold $p^*$ at confidence level $1-\delta$. The results reported here are computed using the SafetyAntVelocity environment and illustrate the evaluation budget required to certify $p^* = 0.05$ at $90\%$ confidence.

\subsection{Zero-Failure Case}

When no failures are observed ($k=0$), the Clopper--Pearson upper bound simplifies approximately to $\hat{p}_{max} \approx 1 - \delta^{1/N}$.


For example, to certify $p^* = 0.05$ at $90\%$ confidence with zero observed failures, approximately $N \approx 45$ episodes suffice.

\subsection{General Case}

When $k > 0$, larger sample sizes are required. Table~\ref{tab:sample_complexity_exact} reports the minimum $N$ required to certify $p^* = 0.05$ at $90\%$ confidence for varying observed failure counts.

\begin{table}[htbp]
\caption{Minimum sample size $N$ required to certify $p^* = 0.05$ at $90\%$ confidence using Clopper--Pearson bounds (SafetyAntVelocity).}
\centering
\begin{tabular}{|p{1.2cm}|c|c|c|}
\hline
Observed Failures $k$ & Minimum $N$ & CP Upper Bound & Empirical $\hat{p}=k/N$ \\
\hline
0   & 45    & 0.04988 & 0.00000 \\
1   & 77    & 0.04958 & 0.01299 \\
2   & 105   & 0.04989 & 0.01905 \\
3   & 132   & 0.04991 & 0.02273 \\
5   & 184   & 0.04982 & 0.02717 \\
10  & 306   & 0.04990 & 0.03268 \\
20  & 538   & 0.04994 & 0.03718 \\
50  & 1202  & 0.04998 & 0.04160 \\
100 & 2275  & 0.04999 & 0.04396 \\
\hline
\end{tabular}
\label{tab:sample_complexity_exact}
\end{table}

As the empirical failure rate approaches the deployment threshold ($\hat{p} \to p^*$), the required evaluation budget grows rapidly. This reflects a fundamental statistical phenomenon: distinguishing a near-threshold policy from an unacceptable one requires increasing statistical evidence.

These calculations provide practical deployment guidance. Rather than fixing a large evaluation budget (e.g., $N=10{,}000$), certification can be performed adaptively by increasing sampling only when preliminary results indicate that certification remains feasible. These values are environment-agnostic but reported here using the SafetyAntVelocity experiment for concreteness.


\end{document}